\documentclass{article} 
\usepackage[T1]{fontenc}
\usepackage{iclr2027_conference,times}

\usepackage{amsmath,amsfonts,bm}

\def\eqref#1{equation~\ref{#1}}

\def\1{\bm{1}}

\DeclareMathAlphabet{\mathsfit}{\encodingdefault}{\sfdefault}{m}{sl}
\SetMathAlphabet{\mathsfit}{bold}{\encodingdefault}{\sfdefault}{bx}{n}

\usepackage{hyperref}
\usepackage{url}
\usepackage{booktabs}
\usepackage{multirow}
\usepackage{graphicx}
\usepackage{xspace}
\usepackage{amssymb}
\usepackage{xcolor}
\usepackage{placeins}
\usepackage{array}
\usepackage{tcolorbox}
\usepackage{enumitem}
\tcbuselibrary{skins,breakable}
\usepackage{authblk}
\usepackage{fontawesome}

\newtcolorbox{promptbox}[1]{%
  enhanced,breakable,
  colback=black!2,colframe=black!35,
  colbacktitle=black!78,coltitle=white,coltext=black,
  title={#1},fonttitle=\small\bfseries,
  fontupper=\small\ttfamily,
  boxrule=0.45pt,arc=2pt,outer arc=2pt,
  boxsep=0pt,left=10pt,right=10pt,top=9pt,bottom=9pt,
  toptitle=5pt,bottomtitle=5pt,
  before skip=12pt,after skip=12pt,
  before upper={\raggedright\setlength{\parindent}{0pt}\setlength{\parskip}{5pt}},
  title after break={#1\hfill\normalfont\small (continued)}
}

\newcommand{\score}{VIEScore2}
\title{\score{}: Unified Image Evaluation with Spatially Grounded Explanations}

\author{
Xianda Du$^{*\spadesuit}$ \quad
Max Ku$^{*\spadesuit}$ \quad
Weiming Ren$^{\spadesuit}$ \quad
Zhi Rui Tam$^{\heartsuit}$ \quad
Chunlin Ren$^{\diamondsuit}$ \quad 
Ping Nie$^{\spadesuit}$ \quad
Min-Hung Chen$^{\clubsuit}$ \quad
Wenhu Chen$^{\spadesuit}$ \\
$^{\spadesuit}$University of Waterloo \quad
$^{\clubsuit}$NVIDIA \quad
$^{\heartsuit}$National Taiwan University \quad
$^{\diamondsuit}$Nanyang Technological University \\
}

\iclrfinalcopy 
\begin{document}

\maketitle

\begin{figure*}[h]
\centering
\includegraphics[width=0.99\textwidth]{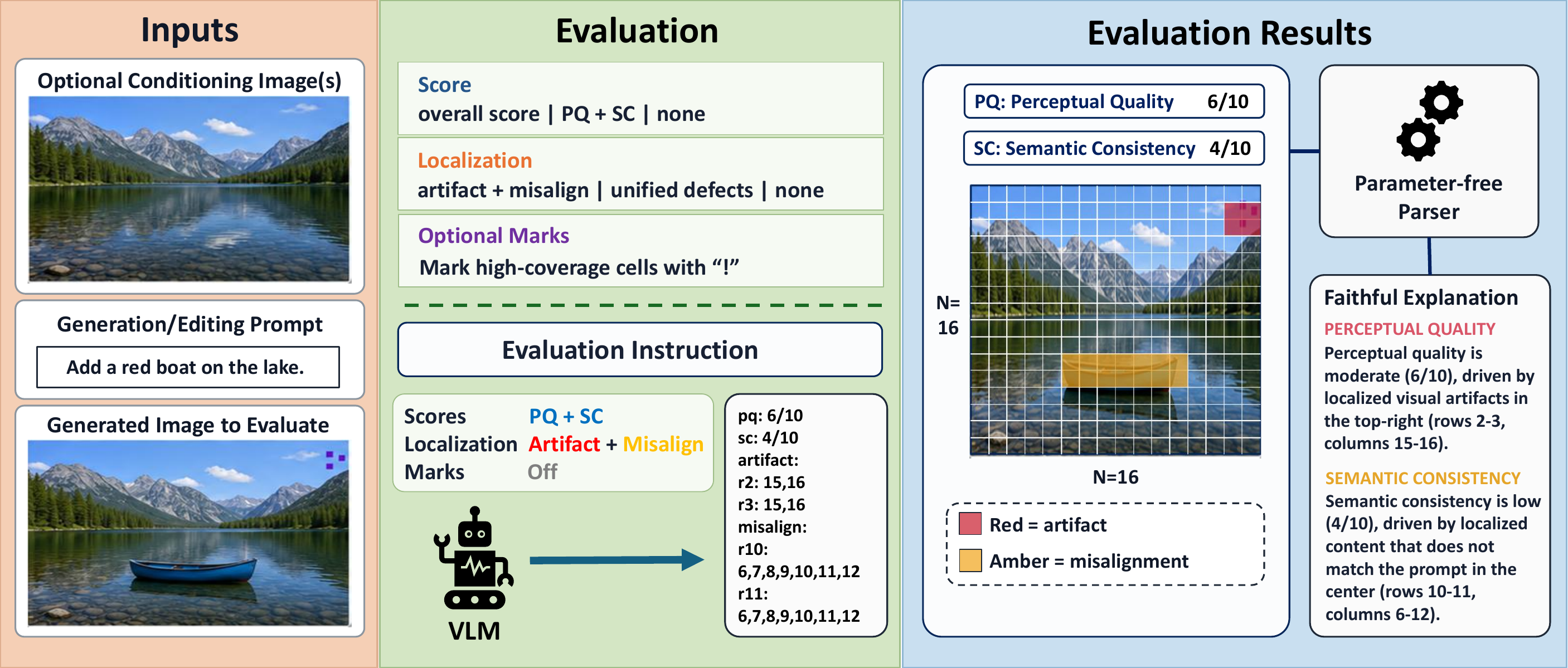}
\caption{An evaluation example with \score{}. The model predicts perceptual quality (PQ), semantic consistency (SC), and a $16\times16$ defect grid, where red and amber denote visual artifacts and semantic misalignments. A parameter-free parser converts the predictions into natural language.}
\label{fig:application}
\end{figure*}

\begin{abstract}
Existing synthetic image evaluators typically provide only a scalar quality score and do not identify the image regions that support it. We introduce \textbf{\score{}}, a unified evaluator for image generation and editing tasks with optional conditioning images. \score{} represents an image as an $N\times N$ grid and jointly predicts quality scores and defect locations in a single model pass. Its text-native grid representation provides a common interface for heterogeneous spatial supervision and enables directly verifiable post-training objectives. We train on 38K examples spanning score-only, localization-only, and joint supervision across generation and editing tasks. Starting from supervised fine-tuning, we further apply GRPO to improve defect localization using rewards that combine cell-level Dice overlap, score accuracy, and output-format validity. A parameter-free parser converts the structured predictions into readable explanations. On the primary suite, \score{} achieves an overall-score SRCC of 0.601, compared with 0.491 for Gemini-3-Flash, the strongest zero-shot general-purpose VLM baseline under matched inputs. For defect localization, \score{} outperforms both general-purpose VLMs and specialized spatial evaluators on three of six benchmarks in per-image grid IoU and ranks among the top three on five, including datasets beyond its training sources.
\end{abstract}

\section{Introduction}

Recent image generation and editing models have achieved increasingly realistic visual synthesis~\citep{rombach2022high,saharia2022photorealistic}. Despite this remarkable progress, generated images still frequently contain localized failures, such as implausible structures, object distortions, visual artifacts, or violations of the input prompt~\citep{liang2024rich,zhang2023perceptual}. Manually identifying these defects is costly, inefficient, and difficult to scale~\citep{ku2024imagenhub}. Consequently, reliable automatic evaluators have become increasingly important. Yet, what should an evaluator produce beyond a single quality score? Humans can often recognize image quality intuitively, but automatic evaluators still struggle to explain their scores and identify defective regions. Most learned evaluators return a scalar score~\citep{hessel2021clipscore,xu2023imagereward,fu2023dreamsim}. VIEScore~\citep{ku2024viescore} prompts a frozen vision-language model (VLM) to produce scores and free-form explanations, but these explanations are not necessarily grounded in image regions. Spatial methods instead predict boxes, heatmaps, or masks~\citep{zhang2026and,guo2025imagedoctor,zhang2023perceptual}, but many focus on text-to-image generation. Extending these methods to image editing and generation tasks that require conditioning images, quality scores and defect categories remains challenging. 

\begin{figure*}[t]
\centering
\includegraphics[width=0.98\textwidth]{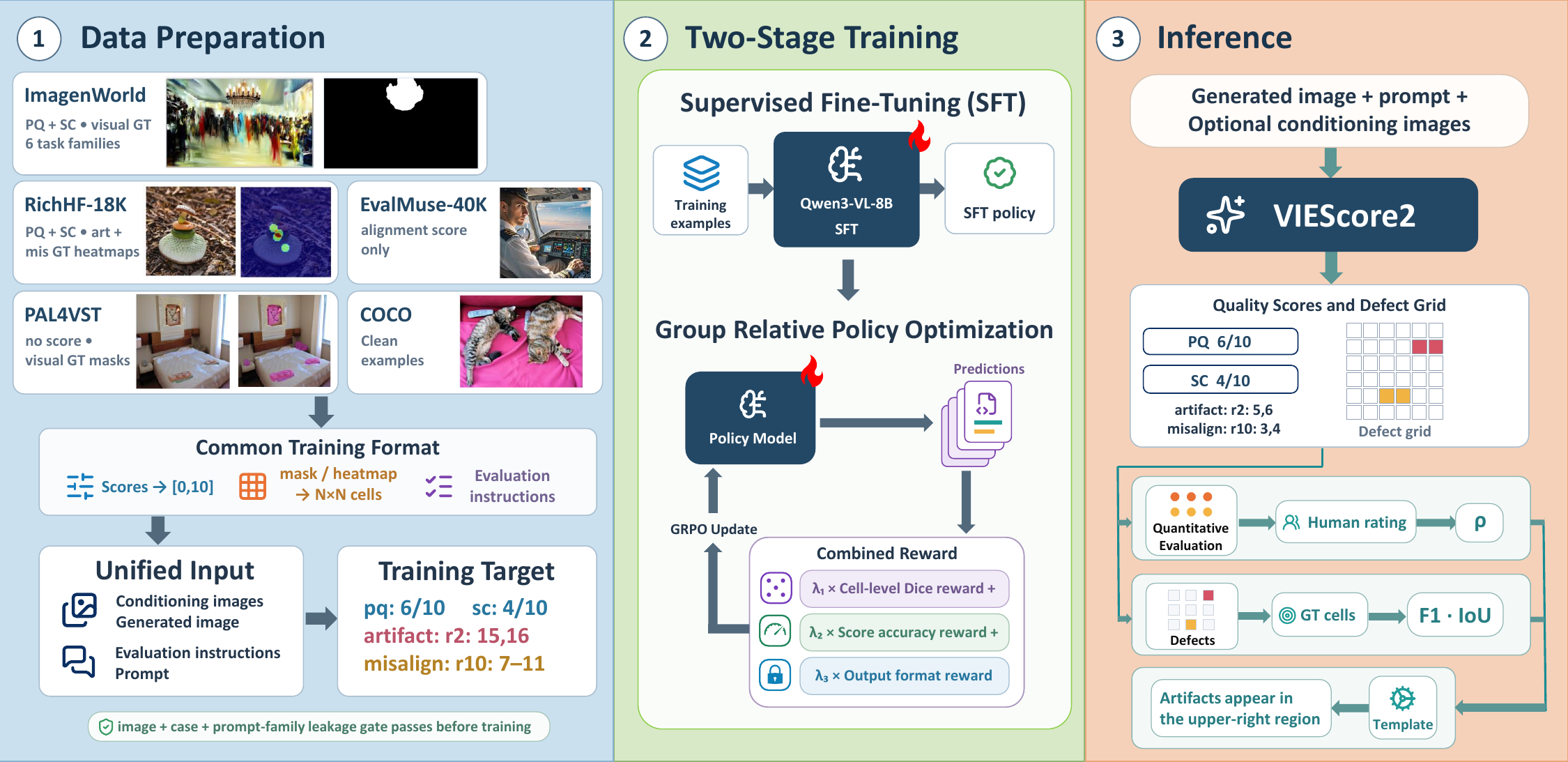}
\caption{Overview of \score{}: data preparation, SFT and GRPO training, and inference. GRPO combines Dice, score, and format rewards. The parameter-free parser converts predicted scores and defect grids into explanations. GT denotes ground truth.}
\label{fig:method}
\end{figure*}

We introduce \score{}, a unified evaluator for diverse image generation and editing tasks. It takes a generated image, its prompt, optional conditioning images, and an evaluation instruction that specifies the requested output fields. \score{} supports score-only, localization-only, and joint evaluation with the same model. It represents defects as sparse cells on an $N\times N$ grid and can separately localize two defect categories: visual artifacts and semantic misalignments. The predicted scores, defect categories, and cell locations are further converted into readable explanations that summarize the scores and spatial evidence. Figure~\ref{fig:application} provides a complete evaluation example. Methodologically, we first apply supervised fine-tuning (SFT) to train a vision-language model (VLM) to predict the quality scores and defect grid. However, its token-level objective does not directly optimize the overlap between predicted and ground-truth cells. We therefore post-train the model using group relative policy optimization (GRPO)~\citep{shao2024deepseekmathpushinglimitsmathematical}. For each input, GRPO samples multiple responses and reinforces those with higher rewards. Our reward combines cell-level Dice overlap with ground-truth defects~\citep{7785132}, score accuracy, and output format validity. We unify data from five sources to train a single evaluator across image generation and editing tasks. The training data include both quality scores and defect locations, although individual examples may provide only one type of supervision. Section~\ref{sec:sft} details the fine-tuning process. Figure~\ref{fig:method} summarizes the data preparation, training, and inference pipeline of \score{}.  Our contributions are threefold:

\begin{enumerate}[noitemsep, topsep=2pt, leftmargin=2em]

\item \textbf{Unified scoring, localization, and explanation.}
We introduce a single evaluator for image generation and editing tasks with zero, one, or multiple conditioning images. Given a generated image, its prompt, optional conditioning images, and an evaluation instruction, \score{} produces quality scores and defect locations without task-specific prediction heads. A parameter-free parser converts these predictions into readable, spatially grounded feedback.

\item \textbf{Verifiable spatial post-training.}
Defects are represented as sparse cells on an image grid. This representation is directly parseable and supports a coverage-weighted cell-level Dice reward that directly rewards localization overlap.

\item \textbf{Learning and evaluation across heterogeneous sources.}
We normalize the scores and spatial annotations from five data sources into a common training format while retaining only the supervision available for each example. Across generation and editing tasks, \score{} achieves higher aggregate overall-score correlation than all four general-purpose VLM baselines under the same input setting, alongside competitive defect localization.

\end{enumerate}

\section{Related Work}


\textbf{Vision-Language Models as Image Evaluators.} 
Recent works have explored using VLMs as automatic image evaluators. Early methods such as TIFA~\citep{hu2023tifaaccurateinterpretabletexttoimage} and DSG~\citep{cho2024davidsonianscenegraphimproving} decompose prompts into structured questions to measure text--image alignment. VQAScore~\citep{lin2024vqascore} uses the probability of an affirmative VQA response as an alignment score, while Q-Align~\citep{wu2023qalign} teaches a VLM to predict text-defined quality levels. EvalMuse-40K~\citep{han2024evalmuse} introduces FGA-BLIP2, an evaluator that predicts overall and fine-grained text--image alignment scores. VIEScore~\citep{ku2024viescore} and its agentic extension CIGEval~\citep{wang2025unifiedagenticframeworkevaluating} further demonstrate that general-purpose VLMs can provide holistic evaluations across multiple conditional image synthesis tasks. These methods primarily return scalar or textual judgments and do not explicitly ground their evaluations in defective image regions. In contrast, \score{} jointly predicts quality scores and spatial defect locations within the same VLM, while supporting generation and editing tasks with zero, one, or multiple conditioning images.

\textbf{Visual Explanation through Defect Localization.}
Spatial evaluators identify defective image regions. PAL~\citep{zhang2023perceptual} predicts pixel-level artifact masks, while RAHF~\citep{liang2024rich} predicts quality scores and separate artifact and misalignment heatmaps. LEGION~\citep{kang2025legion} combines synthetic-image detection, artifact segmentation, and explanation. HEIE~\citep{yang2025heiemllmbasedhierarchicalexplainable} combines implausibility scores, heatmaps, and explanations. These approaches provide spatial feedback, but typically use dedicated spatial outputs such as masks or heatmaps and are mainly designed for text-to-image evaluation. \score{} instead represents defects as sparse grid cells generated directly as text, enabling a single model to produce scores and spatial evidence without task-specific prediction heads. It also supports conditional generation and editing with optional conditioning images, including multiple references. A parameter-free parser converts the structured predictions into readable descriptions without additional training. Appendix~\ref{app:related} compares their output interfaces.

\textbf{Reinforcement Learning for Image Evaluation.}
Reinforcement learning has been used to improve image-quality scoring and reasoning~\citep{li2025qinsight,wu2025visualquality,unifiedreward2025} and, with IoU-based verifiable rewards, visual grounding~\citep{liu2025visualrftvisualreinforcementfinetuning}. ImageDoctor~\citep{guo2025imagedoctor} is the closest to our work and uses grounding, score, and heatmap rewards for text-to-image evaluation. SDG~\citep{zhang2026and} uses rewards for box localization, description consistency, and importance estimation. These methods demonstrate that verifiable spatial supervision can improve image evaluation, but rely on heatmap- or box-based grounding objectives. \score{} instead optimizes the same sparse textual grid representation used at inference through a cell-level Dice reward, together with score and format rewards. This directly aligns post-training with the model's structured output while jointly optimizing quality scoring and defect localization.
\section{Method}
\label{sec:method}

\subsection{Evaluation Inputs and Outputs}
\label{sec:interface}
The input $x_i$ contains a generated image, a generation or editing prompt, and zero, one, or multiple conditioning images. Conditioning images can be source images for editing or reference images for generation. They precede the generated image in the model input. An evaluation instruction $\tau_i$ specifies the requested output fields: \textbf{1. scores}: an overall score or separate scores for perceptual quality (PQ) and semantic consistency (SC), all on a $0$--$10$ scale; \textbf{2. localization}: defect cells on an $N\times N$ grid, either in a single grid or in separate grids for visual artifacts and semantic misalignments; and \textbf{3. coverage marks}: optional \texttt{!} marks on defect cells, as defined in Section~\ref{sec:grid}. The instruction can request scores, localization, or both, and specifies whether to include coverage marks. During training, the requested fields match the available supervision. \score{} generates structured predictions in a single autoregressive pass. The scores and cell coordinates can be parsed directly for reward computation and evaluation. The same VLM supports these output configurations without task-specific prediction heads or additional fine-tuning for each configuration.

\subsection{Sparse Grid Representation}
\label{sec:grid}

We divide the generated image into an $N\times N$ grid and represent defects as a set of cell coordinates $G \subseteq \{1,\ldots,N\}\times\{1,\ldots,N\}$. We encode these coordinates as sparse text sequences for autoregressive generation~\citep{lan2025text4segreimaginingimagesegmentation}. This format accommodates disconnected defect regions and separate defect categories. An empty set $G=\varnothing$ indicates no defects. We use $N=16$ by default to balance localization accuracy and output length. Details of the resolution ablation are provided in Section~\ref{sec:ablations}.

\textbf{Grid construction.} The spatial annotations in the training data are masks or heatmaps. We use resizing and thresholding to convert them into grid cells. For sources that annotate the two defect categories separately, we retain separate cell sets for visual artifacts ($G_{\mathrm{art}}$) and semantic misalignments ($G_{\mathrm{mis}}$). We evaluate each set separately and use their union, $G=G_{\mathrm{art}}\cup G_{\mathrm{mis}}$, for aggregate localization. Sources without separate defect categories use a single grid. For heatmap annotations, we mark defect cells whose resized original heatmap value is at least $0.9$. We take the union of marked cells across defect categories to form $M\subseteq G$. These coverage marks reflect annotation support, not defect severity, and are used to weight the cell-level Dice reward (Section~\ref{sec:rlvr}).

\textbf{Text output.}
The model lists defect cells by row using 1-based indices. For example, \texttt{r5: 8,9!} identifies columns 8 and 9 in row 5 and marks column 9 as having high annotation coverage.
It omits rows without defects and outputs \texttt{none} for an empty
grid. The header lines \texttt{artifact:} and \texttt{misalign:} distinguish the two defect categories when separate grids are requested. This format preserves all defect-cell coordinates without listing
all $N^2$ cells.

\subsection{Supervised Fine-Tuning}
\label{sec:sft}

We combine the five data sources, ImagenWorld~\citep{sani2026imagenworld}, RichHF-18K~\citep{liang2024rich}, EvalMuse-40K~\citep{han2024evalmuse}, PAL4VST~\citep{zhang2023perceptual}, and COCO~\citep{lin2014microsoft}, into a common training format. The resulting training corpus contains around 38K examples spanning score-only, localization-only, and joint supervision, including both unconditional and conditional generation/editing settings. We normalize scores to the shared scale and convert spatial annotations into defect grids as described in Section~\ref{sec:grid}. Each example includes an evaluation instruction and a target response containing only the available supervision. ImagenWorld provides PQ and SC supervision and defect masks across generation and editing tasks, including tasks with conditioning images. EvalMuse's alignment score serves as the overall-score target, PAL4VST provides defect locations, and RichHF provides PQ and SC scores with separate artifact and misalignment grids and coverage marks. COCO images provide defect-free examples.
Appendix~\ref{app:data-eval} details source mappings, grid construction, and data separation. We fine-tune Qwen3-VL-8B~\citep{bai2025qwen3vl} to generate these
target responses using autoregressive cross-entropy as illustrated in Equation~\ref{eq:sft}:
\begin{equation}
\mathcal L_{\mathrm{SFT}}
=
-\sum_i \sum_{t=1}^{|y_i|}
\log p_\theta
\left(y_{i,t}\mid x_i,\tau_i,y_{i,<t}\right),
\label{eq:sft}
\end{equation}
where $\theta$ denotes the trainable model parameters, $y_i$ is the target response, and $y_{i,t}$ is its $t$-th token.
The model predicts each token from the input $x_i$, the evaluation
instruction $\tau_i$, and the preceding target tokens $y_{i,<t}$.
Only target-response tokens contribute to the loss.
Missing supervision is distinct from an empty grid: score-only examples omit the grid field and are never trained to output \texttt{none} for localization. We cap clean examples with localization supervision at $10\%$ of the training corpus to limit the frequency of empty-grid targets.

\subsection{Reinforcement Learning with GRPO}
\label{sec:rlvr}

Starting from the SFT checkpoint, we apply GRPO~\citep{shao2024deepseekmathpushinglimitsmathematical} to optimize the rewards in Eq.~\ref{eq:total-reward}. For each input, the model samples four responses. We parse each response into scores and defect cells and compute its reward against the available ground truth. GRPO normalizes the rewards within each group and uses them to update the model with a clipped policy objective. This encourages responses with higher rewards relative to others for the same input. Throughout the paper, \score{} denotes the final model after GRPO; \score{} ($-\mathrm{GRPO}$) denotes the SFT checkpoint without GRPO. Parenthetical minus signs indicate omitted training stages or reward components. The total reward combines defect localization, score accuracy, and output format validity:
\begin{equation}
r_{\mathrm{total}}
=
\lambda_{\mathrm{dice}}r_{\mathrm{dice}}
+
\lambda_{\mathrm{score}}r_{\mathrm{score}}
+
\lambda_{\mathrm{format}}r_{\mathrm{format}},
\label{eq:total-reward}
\end{equation}
where we use $(\lambda_{\mathrm{dice}},\lambda_{\mathrm{score}},\lambda_{\mathrm{format}})=(1,0.3,0.1)$. All rewards are computed from the parsed outputs without a learned reward model.

\textbf{Cell-level Dice reward.}
We adapt the Dice-based objective~\citep{7785132} to measure overlap between the ground-truth defect set $G$ and its prediction $\hat G$. For separate artifact and misalignment grids, both sets are unions across the two categories, as in Section~\ref{sec:grid}. This reward measures defect localization but does not penalize category swaps. For each cell $c$, we set $w_c=2$ if $c\in M$ and $w_c=1$ otherwise, where $M$ is the union of ground-truth coverage marks. We count correctly predicted and missed defect cells as $\mathrm{TP}_w=\sum_{c\in G\cap\hat G}w_c$ and $\mathrm{FN}_w=\sum_{c\in G\setminus\hat G}w_c$. False positives remain unweighted: $\mathrm{FP}=|\hat G\setminus G|$. The parameter $\beta$ controls the precision--recall trade-off:

\begin{equation}
r_{\mathrm{dice}}
=
\frac{(1+\beta^2)\mathrm{TP}_w}
{(1+\beta^2)\mathrm{TP}_w+\beta^2\mathrm{FN}_w+\mathrm{FP}}.
\label{eq:dice-reward}
\end{equation}
Larger $\beta$ places more emphasis on missed defects (false negatives), while smaller $\beta$ places more emphasis on incorrectly identified defects (false positives). We use $\beta=1$ because it achieves the highest development grid IoU (Section~\ref{sec:frontier}). At this setting, the reward is a coverage-weighted Dice score. With unit cell weights, it reduces to standard Dice, equivalently cell-level $F_1$.

The Dice reward is $1$ when both grids are empty and $0$ when the predicted grid cannot be parsed. Reward computation and evaluation use the same rules to parse defect grids.

\textbf{Score and format rewards.}
For a ground-truth score $s\in[0,10]$ and its prediction $\hat s$, we use $r_{\mathrm{score}}=1-|\hat s-s|/10$ as its score reward. If both PQ and SC are supervised, we average their individual score rewards. A missing or unparseable score receives zero reward. For examples with localization supervision, the format reward $r_{\mathrm{format}}$ is $1$ if the grid can be parsed. For score-only examples, it is $1$ if a score can be parsed. Otherwise, it is $0$. If an example lacks score or localization supervision, we set the corresponding reward term to zero for every response in the group. That term contributes no within-group reward difference.

\subsection{Parameter-Free Parser}
\label{sec:verbalize}

To make \score{}'s predictions easy to read, a parameter-free parser converts the structured output into a textual explanation using fixed rules and templates. It extracts the predicted scores and defect grids, groups connected defect cells, and describes their locations. This step has no trainable parameters. When defect categories are predicted separately, it associates visual artifacts with perceptual quality and semantic misalignments with semantic consistency. The templates use only the predicted scores, defect categories, and locations, keeping explanations faithful to the structured predictions by construction. Appendix~\ref{app:impl-details} provides training and inference settings and the rules used by the parameter-free parser.

\begin{table*}[t]
\centering
\small
\setlength{\tabcolsep}{1.4pt}
\caption{Defect localization on six test sets using a shared $16\times16$ grid. Methods use their supported inputs. IoU averages per-image overlap; $F_1$ pools cell counts. $^{\dagger}$API models use fixed prompts and temperature $0$; $^{\S}$LEGION uses its released intermediate checkpoint.}
\label{tab:spatialcompare}
\renewcommand{\arraystretch}{1.10}
\begin{tabular}{@{}lcccccccccccc@{}}
\toprule
& \multicolumn{4}{c}{Training sources} & \multicolumn{8}{c}{Additional datasets} \\
\cmidrule(lr){2-5}\cmidrule(lr){6-13}
& \multicolumn{2}{c}{RichHF} & \multicolumn{2}{c}{PAL4VST} & \multicolumn{2}{c}{AbHuman} & \multicolumn{2}{c}{HAD} & \multicolumn{2}{c}{SynthScars} & \multicolumn{2}{c}{SDG-30K} \\
\cmidrule(lr){2-3}\cmidrule(lr){4-5}\cmidrule(lr){6-7}\cmidrule(lr){8-9}\cmidrule(lr){10-11}\cmidrule(lr){12-13}
Method & IoU & $F_1$ & IoU & $F_1$ & IoU & $F_1$ & IoU & $F_1$ & IoU & $F_1$ & IoU & $F_1$ \\
\midrule
Qwen3-VL-8B & 0.022 & 0.137 & 0.026 & 0.124 & 0.024 & 0.108 & 0.044 & 0.111 & 0.026 & 0.117 & \underline{0.114} & 0.269 \\
GPT-5.6-terra$^{\dagger}$ & 0.096 & 0.218 & 0.118 & 0.239 & 0.171 & 0.217 & 0.148 & 0.251 & 0.162 & 0.281 & 0.058 & 0.217 \\
GPT-5.6-sol$^{\dagger}$ & 0.156 & 0.274 & 0.141 & 0.248 & \underline{0.217} & 0.282 & \textbf{0.211} & 0.298 & 0.186 & 0.290 & 0.078 & 0.258 \\
Gemini-3-Flash$^{\dagger}$ & 0.047 & 0.082 & 0.048 & 0.068 & 0.057 & 0.079 & 0.075 & 0.119 & 0.062 & 0.101 & 0.031 & 0.068 \\
Claude Opus 5.5$^{\dagger}$ & 0.167 & 0.285 & 0.140 & 0.225 & 0.146 & \textbf{0.308} & 0.142 & 0.235 & 0.196 & 0.358 & 0.084 & 0.216 \\
PAL & 0.014 & 0.035 & \underline{0.306} & \textbf{0.535} & 0.009 & 0.016 & 0.009 & 0.038 & 0.027 & 0.046 & 0.007 & 0.014 \\
SegFormer-b0 & 0.274 & 0.450 & 0.146 & 0.237 & 0.130 & 0.246 & 0.156 & 0.286 & 0.170 & 0.283 & 0.103 & 0.260 \\
RAHF & \underline{0.285} & \textbf{0.474} & 0.105 & 0.169 & 0.141 & 0.260 & 0.164 & 0.281 & 0.172 & 0.294 & 0.068 & 0.164 \\
ImageDoctor & 0.284 & \underline{0.471} & 0.100 & 0.167 & 0.144 & \underline{0.293} & 0.158 & 0.275 & 0.169 & 0.305 & 0.086 & 0.207 \\
LEGION$^{\S}$ & 0.086 & 0.176 & 0.074 & 0.130 & 0.174 & 0.152 & 0.191 & \textbf{0.324} & 0.218 & \underline{0.363} & 0.038 & 0.120 \\
SDG & 0.140 & 0.243 & 0.115 & 0.196 & \textbf{0.253} & 0.292 & 0.154 & 0.228 & \underline{0.222} & 0.330 & \textbf{0.137} & \textbf{0.353} \\
\midrule
\score{} & \textbf{0.299} & 0.466 & \textbf{0.335} & \underline{0.439} & 0.215 & 0.194 & \underline{0.197} & \underline{0.314} & \textbf{0.234} & \textbf{0.368} & 0.102 & \underline{0.272} \\
\bottomrule
\end{tabular}
\end{table*}

\section{Experiments}
\label{sec:experiments}
We assess \score{} on quality scoring and defect localization across image generation and editing tasks and compare it with existing evaluators. We also examine the effects of GRPO and the trade-offs of the sparse grid representation.

\subsection{Evaluation Setup}
\label{sec:setup}

\textbf{Evaluation suites.}
Our primary evaluation suite contains $1{,}300$ held-out examples from the five data sources described in Section~\ref{sec:sft}. Each metric uses only examples with the corresponding ground truth. We evaluate localization on $1{,}100$ examples ($\mathcal E_G$), overall scores on $900$ examples, and PQ and SC scores on $700$ examples. COCO images are included in localization evaluation but excluded from score correlations because their score targets are assigned rather than human-rated. We also evaluate on larger held-out sets from PAL4VST and RichHF and a $2$K EvalMuse subset. To assess transfer beyond the training sources, we use four additional localization datasets: AbHuman~\citep{fang2024humanrefiner}, HAD~\citep{wang2024detecting}, SynthScars~\citep{kang2025legion}, and SDG-30K~\citep{zhang2026and}.

\textbf{Inputs and decoding.}
We distinguish two input settings: $\chi_0$ uses the generated image and prompt, while $\chi_K$ also includes the available conditioning images. The main scoring and joint-evaluation comparisons use $\chi_K$; the $\chi_0$ setting is examined separately.

For each source, \score{} uses a fixed evaluation instruction to specify the requested scores, defect categories, and coverage marks (Section~\ref{sec:interface}). We use greedy decoding and parse the generated text into scores and defect cells.

\textbf{Metrics.}
We evaluate localization on the union of the applicable defect categories. Let $G_i$ and $\hat G_i$ denote the ground-truth and predicted cell sets for image $i$. We pool true-positive, false-positive, and false-negative cell counts across $\mathcal E_G$ to compute micro precision, recall, and $F_1$. We also report mean per-image grid IoU over $\mathcal E_G^+$, the examples with non-empty ground-truth grids:
\begin{equation}
\operatorname{IoU}_p^{\mathrm{grid}}
=
\frac{1}{|\mathcal E_G^+|}
\sum_{i\in\mathcal E_G^+}
\frac{|G_i\cap\hat G_i|}{|G_i\cup\hat G_i|}.
\label{eq:grid-ioup}
\end{equation}
Per-image grid IoU measures localization overlap for each image and gives every image with annotated defects equal weight. This matches our goal of providing useful spatial feedback for individual images, whose defect regions can vary substantially in size. We therefore emphasize per-image grid IoU when comparing localization performance. Micro $F_1$ complements this measure by pooling cell counts across images, giving greater influence to images with larger ground-truth or predicted defect regions. We also report false alarms on clean images. Pixel-IoU is used only for native dense-mask and grid-resolution comparisons. When both PQ and SC are predicted, their geometric mean gives the overall score. For score evaluation, we report Spearman rank correlation (SRCC). Overall, PQ, and SC correlations each use examples with the corresponding score annotations. Correlations use successfully parsed prediction--target pairs and are undefined for constant score vectors. We report valid-pair counts when parsing coverage is incomplete.

\textbf{Compared models.}
We compare \score{} with general-purpose VLMs and specialized image evaluators.
The general-purpose models are Qwen3-VL-8B~\citep{bai2025qwen3vl}, GPT-5.6-terra~\citep{openai2026gpt56terra}, GPT-5.6-sol~\citep{openai2026gpt56sol}, Gemini-3-Flash~\citep{google2025gemini3flash}, and Claude Opus 5.5~\citep{anthropic2026opus55}.
Spatial evaluators include RAHF~\citep{liang2024rich}, ImageDoctor~\citep{guo2025imagedoctor}, PAL~\citep{zhang2023perceptual}, LEGION~\citep{kang2025legion}, and SDG~\citep{zhang2026and}.
We also train SegFormer-b0~\citep{xie2021segformer} on the same spatial supervision as a dense-localization baseline. Each method is evaluated only on the outputs it supports. Appendix~\ref{app:eval-protocol} provides the evaluation subsets, metric conventions, and baseline settings. In tables with ranking annotations, \textbf{bold} denotes the best value and \underline{underlining} the second-best distinct value in each column. Tied values share the same marking.

\subsection{Strong Localization across Benchmarks}
\label{sec:external}

Table~\ref{tab:spatialcompare} compares all spatial predictions on the common grid. The baselines cover heatmaps (RAHF and ImageDoctor), masks (PAL and LEGION), and boxes (SDG). We use their released models and fixed spatial-conversion settings. Each method uses its supported inputs; a shared grid aligns the evaluation space, not the input interfaces.

\score{} ranks first in per-image grid IoU on RichHF ($0.299$), PAL4VST ($0.335$), and SynthScars ($0.234$), second on HAD, and third on AbHuman. It also leads in SynthScars $F_1$ ($0.368$). Other methods lead on different benchmarks: GPT-5.6-sol in HAD IoU, Claude Opus 5.5 in AbHuman $F_1$, and SDG on SDG-30K. The top-three IoU rankings on five benchmarks show competitive localization across training sources and additional datasets. GRPO improves grid IoU over \score{} ($-\mathrm{GRPO}$) on all six datasets and improves $F_1$ on five. Section~\ref{sec:frontier} separates the effect of GRPO from additional SFT and examines the reward components.

\begin{figure}[t]
\centering
\includegraphics[width=0.99\textwidth]{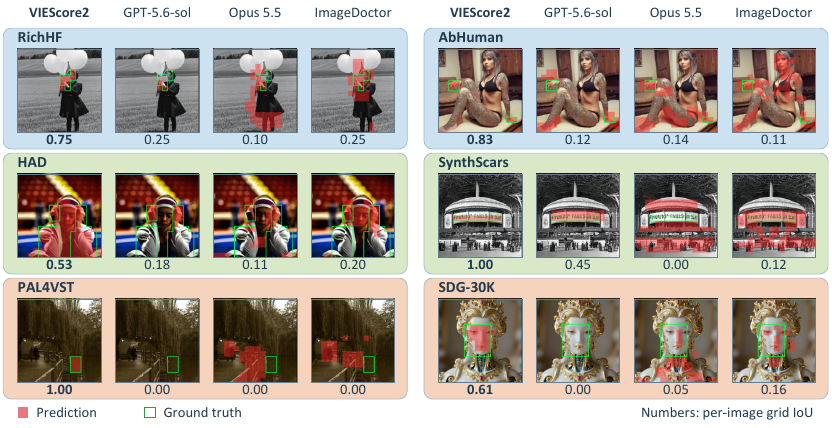}
\caption{
Selected defect-localization examples across six benchmarks. Red cells denote predictions and green outlines denote ground-truth defect regions. Numbers report per-image grid IoU. Additional qualitative examples and failure cases are provided in Appendix~\ref{app:qualitative}.
}
\label{fig:qualgrid}
\end{figure}

\textbf{Qualitative comparison.}
Figure~\ref{fig:qualgrid} compares selected examples from six benchmarks. \score{} more closely matches the annotated regions, including the sign on SynthScars and tree region on PAL4VST. Competing predictions miss these regions or cover surrounding areas. The HAD example also exposes a limitation: \score{} marks unannotated parts of the face.

\subsection{Joint Scoring and Localization across Tasks}
\label{sec:headtohead}
\label{sec:main}

\textbf{Quality scoring.}
Table~\ref{tab:scorecompare} compares overall-score predictions under the same $\chi_K$ input setting. On the primary suite, \score{} achieves the highest aggregate SRCC ($0.601$), followed by Gemini-3-Flash ($0.491$) and GPT-5.6-sol ($0.437$). All five models receive the generated image, prompt, and available conditioning images. ImagenWorld covers text-to-image generation (TIG), text-guided editing (TIE), and generation/editing with one (SRIG/SRIE) or multiple conditioning images (MRIG/MRIE). Each task contains $50$ examples.  \score{} also leads on RichHF, EvalMuse, and SRIG. Gemini-3-Flash leads on TIE and SRIE, GPT-5.6-sol on MRIG and MRIE, and GPT-5.6-terra on TIG.

\begin{table*}[t]
\centering
\small
\setlength{\tabcolsep}{3.0pt}
\caption{Overall-score SRCC on the primary suite ($900$ examples) with matched $\chi_K$ inputs. $^{\dagger}$API correlations use successfully parsed scores. $^{\ddagger}$Qwen3-VL-8B uses our evaluation instructions without fine-tuning.}
\label{tab:scorecompare}
\begin{tabular}{@{}lccccccccc@{}}
\toprule
Method & All & RichHF & EvalMuse & TIG & TIE & SRIG & SRIE & MRIG & MRIE \\
\midrule
Qwen3-VL-8B judge$^{\ddagger}$ & 0.373 & 0.456 & 0.473 & 0.570 & 0.066 & 0.128 & 0.393 & \underline{0.537} & $-0.290$ \\
GPT-5.6-terra$^{\dagger}$ & 0.402 & 0.440 & 0.567 & \textbf{0.640} & 0.493 & 0.258 & 0.110 & 0.523 & 0.539 \\
GPT-5.6-sol$^{\dagger}$ & 0.437 & 0.470 & \underline{0.655} & \underline{0.623} & \underline{0.590} & 0.275 & 0.253 & \textbf{0.539} & \textbf{0.684} \\
Gemini-3-Flash$^{\dagger}$ & \underline{0.491} & \underline{0.515} & 0.535 & 0.531 & \textbf{0.686} & \underline{0.356} & \textbf{0.404} & 0.418 & \underline{0.677} \\
\midrule
\score{} & \textbf{0.601} & \textbf{0.692} & \textbf{0.803} & 0.589 & 0.429 & \textbf{0.375} & \underline{0.403} & 0.509 & 0.417 \\
\bottomrule
\end{tabular}
\end{table*}

\textbf{Joint evaluation.}
Table~\ref{tab:main} compares joint localization and quality scoring under the same $\chi_K$ input setting. \score{} achieves higher localization $F_1$, grid IoU, and PQ and SC correlations than the two baselines evaluated in this joint setting. On the same 250 conditional examples, providing the available conditioning images improves overall-score SRCC from 0.420 to 0.451 and localization F1 from 0.514 to 0.536. This comparison keeps the model, evaluation instruction, decoding, grid-parsing rules, and evaluation support fixed, isolating the contribution of the conditioning inputs. The MMRB2 preference evaluation shows competitive text-to-image performance but weaker editing accuracy than the general-purpose VLM baselines. Appendix~\ref{app:additional}  provides the full task-level results and the separate $\chi_0$ comparisons.

\begin{table}[t]
\centering
\small
\caption{Joint evaluation with matched $\chi_K$ inputs. Localization uses $1{,}100$ examples; PQ/SC SRCC uses $700$. Grid IoU excludes empty ground-truth grids. $^{\dagger}$GPT-5.6-terra returns valid scores for $554/700$ examples.}
\label{tab:main}
\begin{tabular}{lcccc}
\toprule
Model & Grid IoU & $F_1$ & PQ SRCC & SC SRCC \\
\midrule
Qwen3-VL-8B & 0.070 & \underline{0.273} & 0.181 & 0.352 \\
GPT-5.6-terra$^{\dagger}$ & \underline{0.112} & 0.247 & \underline{0.357} & \underline{0.508} \\
\midrule
\score{} & \textbf{0.324} & \textbf{0.506} & \textbf{0.558} & \textbf{0.564} \\
\bottomrule
\end{tabular}
\end{table}

\subsection{GRPO Improves Localization beyond Additional SFT}
\label{sec:frontier}

Across three independent runs, GRPO improves grid IoU by $0.024$--$0.029$ over the shared SFT checkpoint, whereas an additional SFT epoch gives $0.001$. These gains show that additional SFT does not reproduce the localization improvement from GRPO. We compare reward components under matched training examples, sampling budget, and optimization steps. \score{} ($-r_{\mathrm{score}}, -r_{\mathrm{format}}$) uses only the Dice reward and raises grid IoU from $0.296$ to $0.320$, accounting for most of the gain. The full reward reaches 0.324 IoU and 0.506 F1 at an overall-score SRCC of 0.601. The Dice-only variant reaches the same SRCC (Table~\ref{tab:rewarddecomp}), so the score reward does not measurably change scoring on the primary suite; its main effect is a small F1 gain (0.481 to 0.506). The format reward has no measurable effect because every variant already parses on all localization examples. We retain both terms as safeguards for training runs where parse failures or score drift could occur, but the localization gain is attributable to the cell-level Dice reward alone.

The precision--recall ablation supports our choice of $\beta=1$ for grid IoU. On the primary suite, GRPO improves both precision ($0.415$ to $0.466$) and recall ($0.545$ to $0.552$). Appendix~\ref{app:grpo-studies} reports the complete GRPO ablations, seed comparisons, and per-category and clean-image diagnostics. Appendix~\ref{app:qualitative} also shows improved overlap and missed defects after GRPO.

\subsection{Grid Resolution Balances Fidelity and Output Length}
\label{sec:ablations}

\begin{figure}[t]
\centering
\includegraphics[width=0.99\textwidth]{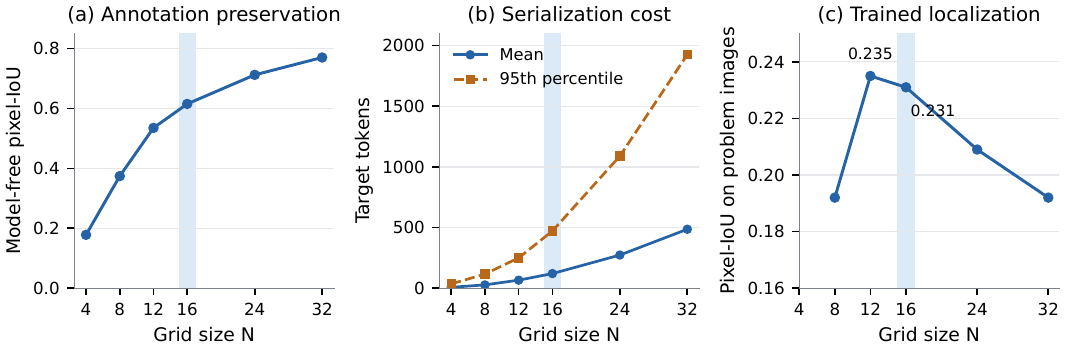}
\caption{Grid resolution versus annotation fidelity, output length, and trained localization. Annotation fidelity compares grids with native masks. Shading marks $N=16$.}
\label{fig:grid-tradeoff}
\label{fig:app-grid-tradeoff}
\end{figure}

Figure~\ref{fig:grid-tradeoff} shows how grid resolution affects annotation fidelity, output length, and trained localization. Trained pixel-IoU$_p$ is similar at $N=12$ and $N=16$ ($0.235$/$0.231$), then falls to $0.192$ at $N=32$. We choose $N=16$ for its higher annotation fidelity than $N=12$ (model-free pixel-IoU $0.615$/$0.535$), while its 95th-percentile target is shorter than at $N=32$ ($471$/$1{,}926$ tokens). This choice balances spatial detail and output length rather than maximizing every metric. Although bounding boxes achieve higher $F_1$ in the matched-budget localization-only ablation ($0.463$ vs. $0.397$), we use sparse cells for their simple text format and direct cell-level reward computation. Appendix~\ref{app:representation} provides the complete representation and resolution comparisons.

\section{Limitations}
\label{sec:limitations}

\textbf{Model choice.}
We validate \score{} only with Qwen3-VL-8B, so its effectiveness across other VLM backbones and model scales remains unknown. In addition, comparisons with released evaluators do not disentangle the effects of architecture, training data, and training procedure.


\textbf{Representation and explanation scope.}
The fixed grid can miss small defects and only coarsely approximate irregular boundaries. The parameter-free parser produces explanations that are faithful to the predicted scores and defect grids, but does not independently verify their correctness or identify defect causes. GRPO can increase false alarms on some clean-image subsets, reflecting a trade-off between stronger defect localization and conservative clean-image recognition. Appendix~\ref{app:ablations} provides the full representation and resolution comparisons.
\section{Conclusion}
We introduced \score{}, a unified framework for explainable image evaluation across generation and editing tasks. Its sparse grid representation unifies heterogeneous spatial supervision across datasets and supports joint prediction of quality scores and defect locations. Using 38K training examples spanning score-only, localization-only, and joint supervision, we train the evaluator through supervised fine-tuning followed by GRPO with verifiable rewards. A parameter-free parser then produces explanations grounded in the structured predictions. Experiments demonstrate stronger alignment with human quality ratings than general-purpose VLMs under matched inputs, together with competitive localization across six benchmarks, including datasets beyond the training sources. Ablations further show that GRPO improves localization beyond additional supervised fine-tuning. 

\clearpage

\section*{Ethics Statement}
\score{} evaluates generated images using existing research datasets; no new human annotations were collected. We use these datasets under their respective licenses. The model may flag legitimate image content as defective or miss actual defects. Its explanations reflect its predictions and do not independently verify them. Scores, defect grids, and explanations should therefore support human judgment rather than replace it.


\bibliography{iclr2027_conference}
\bibliographystyle{iclr2027_conference}

\newpage
\appendix
\section{Datasets and Annotation Mappings}
\label{app:data-eval}

\subsection{Source Supervision and Target Construction}
\label{app:source-mapping}

\paragraph{Source supervision.}
Table~\ref{tab:source-contract} summarizes the score and spatial supervision available from each source.

\begin{table}[!htbp]
\centering
\small
\setlength{\tabcolsep}{3pt}
\renewcommand{\arraystretch}{1.15}
\caption{Supervision and requested output fields by source. ``--'' indicates an omitted field.}
\begin{tabular*}{\textwidth}{@{\extracolsep{\fill}}lccc@{}}
\toprule
Source & Scores & Localization & Raw supervision \\
\midrule
ImagenWorld & PQ, SC & single grid & ratings, masks \\
RichHF & PQ, SC & \shortstack{artifact and\\misalignment grids} & ratings, heatmaps \\
PAL4VST & -- & single grid & masks \\
EvalMuse & overall & -- & alignment ratings \\
COCO & PQ, SC & single grid & clean constraint \\
\bottomrule
\end{tabular*}
\label{tab:source-contract}
\end{table}

\paragraph{Score normalization.}
The source mappings retain the meaning of each available rating. For RichHF, the normalized PQ target is the mean of the artifact and aesthetic ratings, and SC is the normalized misalignment rating. For ImagenWorld, each $1$--$5$ rating $r$ is mapped to $0$--$1$ by $(r-1)/4$; PQ averages the normalized artifact and aesthetic-quality ratings, and SC uses prompt relevance. Both are then scaled to $0$--$10$. EvalMuse's $1$--$5$ mean alignment rating $r$ is mapped to $0$--$10$ by $2.5(r-1)$. Training targets are rounded to integer scores. COCO supplies clean examples with PQ and SC targets of $10$ and empty defect grids. These assigned targets are excluded from score correlations.

\paragraph{Spatial annotation processing.}
\label{app:heatmap-processing}
Binary masks are resized to $N\times N$ using LANCZOS and thresholded at $0.5$. RichHF heatmaps are first binarized at $0.5$ before resizing; the two defect categories are processed separately and combined only for aggregate evaluation. Coverage marks identify defect cells whose resized original heatmap response is at least $0.9$. They measure annotation support rather than severity. Bounding boxes are filled before rasterization.

\begin{figure}[!htbp]
\centering
\includegraphics[width=\textwidth]{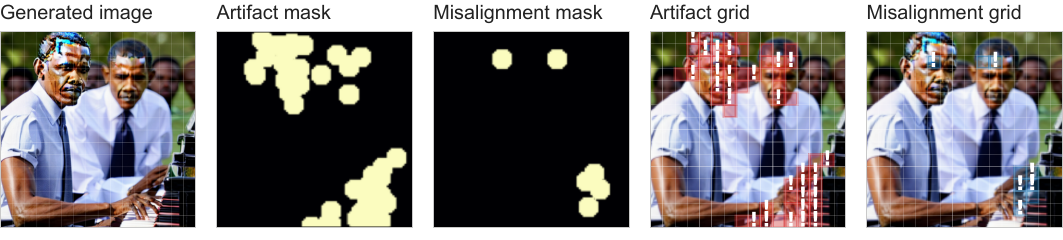}
\caption{RichHF annotations and their $16\times16$ grid targets. Red and blue denote artifact and misalignment; ``!'' marks high annotation coverage.}
\label{fig:app-source-mapping}
\end{figure}

\FloatBarrier
\subsection{Corpus Composition and Data Separation}
\label{app:data-split}

\paragraph{Training corpus.}
Table~\ref{tab:app-data-counts} summarizes the training, validation, and primary evaluation splits. COCO images use human-written captions and provide clean-image supervision.

\begin{table}[!htbp]
\centering
\small
\setlength{\tabcolsep}{3pt}
\renewcommand{\arraystretch}{1.15}
\caption{Data splits by source. Counts are records, which may share an image.}
\begin{tabular*}{\textwidth}{@{\extracolsep{\fill}}lrrrr@{}}
\toprule
Source & Train & Validation & Primary eval. & \shortstack{Conditioning\\images} \\
\midrule
RichHF & 13{,}920 & 1{,}518 & 400 & no \\
EvalMuse & 9{,}000 & 1{,}000 & 200 & no \\
PAL4VST & 8{,}177 & 908 & 300 & no \\
ImagenWorld editing/reference & 3{,}642 & 415 & 250 & yes \\
ImagenWorld TIG & 1{,}165 & 121 & 50 & no \\
COCO & 2{,}218 & 272 & 100 & no \\
\midrule
Total & 38{,}122 & 4{,}234 & 1{,}300 & \\
\bottomrule
\end{tabular*}
\label{tab:app-data-counts}
\end{table}

\paragraph{Training and evaluation separation.}
Training excludes evaluation images, alternate annotations of the same image, and held-out ImagenWorld cases. Exact and perceptual image hashes are checked for overlap.

A prompt-overlap audit changes overall SRCC by less than $0.001$ after removing the affected EvalMuse example. The primary suite shares generators with training; transfer is assessed separately on additional localization datasets.

\FloatBarrier
\section{Implementation Details}
\label{app:impl-details}

\subsection{Optimization and Inference}
\label{app:optimization}

\paragraph{Input processing.}
Generated and conditioning images are resized with preserved aspect ratio to at most $384^2$ and $256^2$ pixels, respectively. The SFT sequence limit is $2{,}048$ tokens.

\paragraph{SFT.}
We fine-tune Qwen3-VL-8B with LoRA~\citep{hu2021loralowrankadaptationlarge} on all linear layers (rank $32$, scaling $16$, dropout $0.05$). AdamW uses a learning rate of $2\times10^{-4}$, cosine decay, $3\%$ warmup, zero weight decay, gradient clipping at $1.0$, and effective batch size $32$. Training runs for five epochs; epoch 4 initializes GRPO, and epoch 5 provides the additional-SFT control. A separate $1\%$ training subset monitors loss.

\paragraph{GRPO.}
The epoch-4 adapter is merged into the backbone for full-parameter post-training. We use 8-bit AdamW with learning rate $10^{-5}$, one epoch, four responses per example, eight gradient-accumulation steps, and no KL penalty. Each run samples $2{,}400$ training examples stratified by supervision type, conditioning, and defect density. Completion length is limited to $768$ tokens. Rewards follow Sec.~\ref{sec:rlvr} with $(\lambda_{\mathrm{dice}},\lambda_{\mathrm{score}},\lambda_{\mathrm{format}})=(1,0.3,0.1)$. The Dice reward is computed on the union of the artifact and misalignment grids. The format reward checks that the grid parses for examples with localization supervision and that the score parses for score-only examples. Unavailable supervision contributes no advantage. Three runs use independent seeds and sampled subsets.

\paragraph{Inference.}
We use greedy decoding with a limit of $1{,}024$ output tokens.

\subsection{Evaluation Instructions and Parameter-Free Parser}
\label{app:output-protocols}
\label{app:renderer}
Each source requests only its supported fields (Table~\ref{tab:source-contract}). EvalMuse has no grid target, whereas a clean localization example uses \texttt{none}. Conditioning images precede the generated image. The full evaluation instructions are given in Appendix~\ref{app:prompts}.

The parameter-free parser groups four-connected cells separately for each defect category and describes up to three components, ordered by size. Each description gives a coarse location and coordinate bounds; additional components are counted. Scores below $5$, from $5$ to below $8$, and at least $8$ are described as low, moderate, and high. Coverage wording depends on the fraction of marked cells. Figure~\ref{fig:app-walkthrough} shows a complete output and its generated explanation. The explanation preserves the prediction's content but does not establish its correctness.

\FloatBarrier
\section{Evaluation Setup and Baselines}
\label{app:eval-protocol}

\subsection{Evaluation Support}
\label{app:eval-support}

\paragraph{Primary evaluation suite.}
Table~\ref{tab:app-eval-support} gives the examples supporting each metric. Scores and localization are evaluated only where the corresponding annotations are available. COCO score targets are assigned and excluded from score correlations; EvalMuse has no localization targets.

\begin{table}[!htbp]
\centering
\small
\setlength{\tabcolsep}{3pt}
\renewcommand{\arraystretch}{1.15}
\caption{Examples available for each metric in the primary suite.}
\begin{tabular*}{\textwidth}{@{\extracolsep{\fill}}lrrrr@{}}
\toprule
Primary source & Total & Localization & Overall score & PQ/SC \\
\midrule
RichHF & 400 & 400 & 400 & 400 \\
PAL4VST & 300 & 300 & 0 & 0 \\
EvalMuse & 200 & 0 & 200 & 0 \\
ImagenWorld (six tasks) & 300 & 300 & 300 & 300 \\
COCO & 100 & 100 & 0 & 0 \\
\midrule
Total & 1{,}300 & 1{,}100 & 900 & 700 \\
\bottomrule
\end{tabular*}
\label{tab:app-eval-support}
\end{table}

\paragraph{Extended evaluation sets.}
Table~\ref{tab:app-external-support} lists the evaluation sets beyond the primary suite. RichHF reserves $100$ images for threshold tuning. The AbHuman~\citep{fang2024humanrefiner} subset uses seed $42$; HAD~\citep{wang2024detecting} includes DALL-E~2/3 ($159/199$ images) and Midjourney/SDXL ($350$ each).

\begin{table}[!htbp]
\centering
\small
\setlength{\tabcolsep}{3pt}
\renewcommand{\arraystretch}{1.15}
\caption{External evaluation subsets. AbHuman, HAD, SynthScars, and SDG-30K report localization on non-empty ground-truth grids.}
\begin{tabular*}{\textwidth}{@{\extracolsep{\fill}}llr@{}}
\toprule
Dataset & Evaluation subset & $n$ \\
\midrule
PAL4VST & Full test set & 1{,}405 \\
RichHF & Test split after threshold tuning & 840 \\
EvalMuse & External score subset & 2{,}000 \\
AbHuman & $2$K validation subset & 1{,}407 \\
HAD & Validation subset & 709 \\
SynthScars & Official test split ($1{,}000$ images) & 951 \\
SDG-30K & Official test split ($1{,}154$ images) & 887 \\
MMRB2 T2I / Edit & Preference pairs & 1{,}000 each \\
\bottomrule
\end{tabular*}
\label{tab:app-external-support}
\end{table}

\subsection{Scores, Empty Grids, and Parse Failures}
\label{app:metric-details}
When a model emits both PQ and SC, its overall prediction is their geometric mean; a score-only evaluator uses its native scalar. Predictions are compared with the source's stored overall target. SRCC uses average ranks for ties, and correlations are undefined for constant vectors. A scorer's valid-pair count must accompany a correlation whenever parsing coverage is incomplete. A same-support comparison additionally restricts both methods to the same valid pairs.

Micro precision, recall, and $F_1$ pool cell counts across the localization support. In contrast, $\operatorname{IoU}^{\mathrm{grid}}_p$ averages per-image IoU only over non-empty ground-truth grids, as in Eq.~\ref{eq:grid-ioup}. A successfully parsed \texttt{none} is an empty prediction. An unparseable grid receives no true positives and misses all positive ground-truth cells; it remains a parse failure and must not be reported as successful clean-image recognition. For clean images, false-alarm counts report how many images have at least one predicted defect cell. They complement $\operatorname{IoU}^{\mathrm{grid}}_p$, which does not measure clean-image behavior.

For the $900$-example score comparison in Table~\ref{tab:scorecompare}, unparseable scores exclude $154$ examples for GPT-5.6-terra and at most $2$ for each other API model. Assigning the scale midpoint to these failures reduces GPT-5.6-terra's aggregate SRCC to $0.357$ without changing the other rows at the reported precision.

\subsection{Baseline Adaptation}
\label{app:baseline-protocol}
Spatial predictions are rasterized into the same $16\times16$ grid when reporting grid metrics. Artifact and misalignment heatmaps are mean-pooled to $16\times16$ and combined as $\max(\text{artifact},\text{misalignment})$; box and mask outputs use the same grid conversion as the ground truth. All released thresholds and tuned settings remain fixed across benchmarks. Table~\ref{tab:app-baseline-settings} lists these settings.

\begin{table}[!htbp]
\centering
\small
\setlength{\tabcolsep}{3pt}
\renewcommand{\arraystretch}{1.15}
\caption{Spatial conversion settings, fixed across benchmarks.}
\begin{tabular*}{\textwidth}{@{\extracolsep{\fill}}llcl@{}}
\toprule
Method & Spatial output & Threshold & Setting source \\
\midrule
RAHF & Two heatmaps & 0.06 & RichHF ($100$) \\
ImageDoctor & Two heatmaps & 0.03 & RichHF ($100$) \\
PAL & Artifact mask & 0.5 & Released default \\
SegFormer-b0 & Defect probabilities & 0.75 & Validation \\
LEGION & Artifact mask & 0.5 cell fraction & Released rule \\
SDG & Defect boxes & -- & Released output \\
\bottomrule
\end{tabular*}
\label{tab:app-baseline-settings}
\end{table}

\FloatBarrier
\subsection{Evaluator Output Interfaces}
\label{app:related}

Table~\ref{tab:app-landscape} summarizes the output interfaces of closely
related evaluators.

\begin{table*}[!htbp]
\centering
\small
\setlength{\tabcolsep}{3pt}
\renewcommand{\arraystretch}{1.15}
\caption{Evaluator output interfaces. Cond. denotes support for conditioning images; ``--'' denotes an unsupported output or training setting.}
\begin{tabular*}{\textwidth}{@{\extracolsep{\fill}}lcclcc@{}}
\toprule
Method & Scores & Spatial output & Text feedback & GRPO & Cond. \\
\midrule
VIEScore~\citep{ku2024viescore} & \checkmark & -- & Rationale & -- & \checkmark \\
RAHF~\citep{liang2024rich} & \checkmark & Heatmaps & -- & -- & -- \\
ImageDoctor~\citep{guo2025imagedoctor} & \checkmark & Heatmap & Reasoning & \checkmark & -- \\
PAL~\citep{zhang2023perceptual} & -- & Mask & -- & -- & -- \\
LEGION~\citep{kang2025legion} & -- & Mask & Explanation & -- & -- \\
FGA-BLIP2~\citep{han2024evalmuse} & \checkmark & -- & -- & -- & -- \\
HEIE~\citep{yang2025heiemllmbasedhierarchicalexplainable} & \checkmark & Heatmap & Analysis & -- & -- \\
SDG~\citep{zhang2026and} & -- & Boxes & Defect text & \checkmark & -- \\
\score{}\ (ours) & \checkmark & $16\!\times\!16$ grid & \shortstack{Parameter-free\\parser} & \checkmark & \checkmark \\
\bottomrule
\end{tabular*}
\label{tab:app-landscape}
\end{table*}

RAHF and ImageDoctor predict four scores; \score{} predicts an overall score or separate PQ and SC scores, as requested. GRPO optimizes scores and heatmaps in ImageDoctor, structured defects in SDG, and the cell-level Dice reward and scores in \score{}.

\FloatBarrier
\section{Additional Evaluation Results}
\label{app:additional}

\subsection{Primary-Suite Results by Metric and Task}
\label{app:primary-results}
Table~\ref{tab:primary-full} expands the joint evaluation in Table~\ref{tab:main} with precision, recall, and \score{} ($-\mathrm{GRPO}$). Relative to SFT, GRPO improves SC correlation but lowers PQ correlation from $0.580$ to $0.558$. Table~\ref{tab:taskwise} reports \score{} results by task and source. Each ImagenWorld task contains $50$ examples, so these task-level comparisons are diagnostic.

\begin{table}[!htbp]
\centering
\small
\setlength{\tabcolsep}{3pt}
\renewcommand{\arraystretch}{1.15}
\caption{Full primary-suite results: localization on $1{,}100$ examples and PQ/SC SRCC on $700$. Rankings compare the $\chi_K$ VLMs; SegFormer uses only the generated image. $^{\dagger}$GPT-5.6-terra returns valid scores for $554/700$ examples.}
\begin{tabular*}{\textwidth}{@{\extracolsep{\fill}}lcccccc@{}}
\toprule
Model & $Precision$ & $Recall$ & $F_1$ & $\operatorname{IoU}^{\mathrm{grid}}_p$
& PQ SRCC & SC SRCC \\
\midrule
Qwen3-VL-8B & 0.211 & 0.388 & 0.273 & 0.070 & 0.181 & 0.352 \\
GPT-5.6-terra$^{\dagger}$ & 0.257 & 0.239 & 0.247 & 0.112 & 0.357 & 0.508 \\
\score{} ($-\mathrm{GRPO}$) & \underline{0.415} & \underline{0.545} & \underline{0.471} & \underline{0.296} & \textbf{0.580} & \underline{0.532} \\
\score{} & \textbf{0.466} & \textbf{0.552} & \textbf{0.506} & \textbf{0.324} & \underline{0.558} & \textbf{0.564} \\
\midrule
SegFormer-b0 (matched loc.) & 0.433 & 0.573 & 0.493 & 0.272 & -- & -- \\
\bottomrule
\end{tabular*}
\label{tab:primary-full}
\end{table}

\begin{table}[!htbp]
\centering
\small
\setlength{\tabcolsep}{3pt}
\renewcommand{\arraystretch}{1.15}
\caption{Primary-suite results by task and source. $\rho_{\mathrm{o}}$ denotes overall-score SRCC; each ImagenWorld task has $50$ examples.}
\begin{tabular*}{\textwidth}{@{\extracolsep{\fill}}lcclcc@{}}
\toprule
Task/source & $F_1$ & $\rho_{\mathrm{o}}$
& Task/source & $F_1$ & $\rho_{\mathrm{o}}$ \\
\midrule
TIG  & 0.572 & 0.589 & MRIG & 0.545 & 0.509 \\
TIE  & 0.692 & 0.429 & MRIE & 0.496 & 0.417 \\
SRIG & 0.512 & 0.375 & RichHF & 0.465 & 0.692 \\
SRIE & 0.412 & 0.403 & PAL4VST & 0.376 & -- \\
     &       &       & EvalMuse & -- & 0.803 \\
\bottomrule
\end{tabular*}
\label{tab:taskwise}
\end{table}

\FloatBarrier
\subsection{Conditioning Images and Score Evaluation}
\label{app:input-ablation}
\label{app:protocol-controls}
\label{app:score-chi0}
Table~\ref{tab:app-score-chi0} reports the $\chi_0$ setting, which uses only the generated image and prompt. These results are separate from the $\chi_K$ comparison in Table~\ref{tab:scorecompare} because access to conditioning images differs.
Score evaluators include ImageReward~\citep{xu2023imagereward}, PickScore~\citep{kirstain2023pickapic}, HPSv2~\citep{wu2023hpsv2}, VQAScore~\citep{lin2024vqascore}, FGA-BLIP2~\citep{han2024evalmuse}, Q-Align~\citep{wu2023qalign}, Q-Insight~\citep{li2025qinsight}, and OmniQuality-R~\citep{lu2025omniquality}.
ImageDoctor~\citep{guo2025imagedoctor} also provides quality scores under this setting. Among these evaluators, FGA-BLIP2 achieves the highest aggregate and EvalMuse SRCC, while ImageDoctor leads on RichHF. Table~\ref{tab:context} isolates the effect of conditioning images using the same \score{} model on the same conditional examples.

\begin{table*}[!htbp]
\centering
\small
\setlength{\tabcolsep}{3pt}
\renewcommand{\arraystretch}{1.15}
\caption{Overall-score SRCC with the generated image and prompt only ($\chi_0$). Rankings compare methods within this input setting.}
\label{tab:app-score-chi0}
\begin{tabular*}{\textwidth}{@{\extracolsep{\fill}}lccccccccc@{}}
\toprule
Method & All & RichHF & EvalMuse & TIG & TIE & SRIG & SRIE & MRIG & MRIE \\
\midrule
ImageReward & 0.239 & 0.227 & 0.595 & 0.107 & $-0.278$ & 0.195 & 0.151 & 0.035 & 0.263 \\
PickScore & 0.233 & 0.263 & 0.441 & 0.412 & $-0.143$ & \underline{0.288} & 0.184 & 0.391 & \textbf{0.407} \\
HPSv2 & 0.166 & 0.182 & 0.473 & 0.209 & 0.034 & 0.181 & 0.177 & \textbf{0.486} & \underline{0.406} \\
VQAScore & 0.224 & 0.238 & 0.440 & 0.110 & 0.081 & 0.145 & 0.085 & 0.105 & 0.140 \\
FGA-BLIP2 & \textbf{0.423} & \underline{0.452} & \textbf{0.918} & \underline{0.422} & $-0.141$ & 0.255 & 0.151 & $-0.062$ & 0.233 \\
ImageDoctor & \underline{0.411} & \textbf{0.714} & 0.280 & \textbf{0.459} & 0.160 & 0.271 & \textbf{0.355} & \underline{0.479} & 0.299 \\
Q-Align & 0.095 & 0.081 & 0.274 & 0.113 & \textbf{0.231} & 0.142 & 0.238 & 0.320 & 0.181 \\
Q-Insight & 0.054 & 0.090 & 0.253 & 0.037 & $-0.011$ & $-0.106$ & \underline{0.243} & 0.086 & $-0.102$ \\
OmniQuality-R & 0.343 & 0.371 & \underline{0.744} & 0.325 & \underline{0.167} & \textbf{0.342} & 0.132 & 0.265 & 0.275 \\
\bottomrule
\end{tabular*}
\end{table*}

\paragraph{Conditioning images.}
\label{app:context}
Table~\ref{tab:context} compares the same $250$ conditional examples with and without conditioning images, keeping the model, evaluation instruction, decoding, grid-parsing rules, and evaluation support fixed. Aggregate gains are $0.030$ in overall-score SRCC and $0.022$ in $F_1$, computed before rounding.

\begin{table}[!htbp]
\centering
\small
\setlength{\tabcolsep}{3pt}
\renewcommand{\arraystretch}{1.15}
\caption{Paired comparison with ($\chi_K$) and without ($\chi_0$) conditioning images. Changes are computed before rounding.}
\begin{tabular*}{\textwidth}{@{\extracolsep{\fill}}lrrrrrr@{}}
\toprule
Task & $n$ & $\rho_{\mathrm{o}}[\chi_K]$ & $\rho_{\mathrm{o}}[\chi_0]$ &
$F_1[\chi_K]$ & $F_1[\chi_0]$ & $\Delta F_1$ \\
\midrule
TIE  & 50 & 0.429 & 0.403 & 0.692 & 0.683 & $+0.009$ \\
SRIG & 50 & 0.375 & 0.378 & 0.512 & 0.503 & $+0.009$ \\
SRIE & 50 & 0.403 & 0.265 & 0.412 & 0.370 & $+0.042$ \\
MRIG & 50 & 0.509 & 0.470 & 0.545 & 0.522 & $+0.023$ \\
MRIE & 50 & 0.417 & 0.509 & 0.496 & 0.458 & $+0.039$ \\
\midrule
All & 250 & 0.451 & 0.420 & 0.536 & 0.514 & $+0.022$ \\
\bottomrule
\end{tabular*}
\label{tab:context}
\end{table}

\FloatBarrier
\subsection{EvalMuse Subset Analysis}
\label{app:evalmuse-results}

\begin{figure}[!htbp]
\centering
\includegraphics[width=\textwidth]{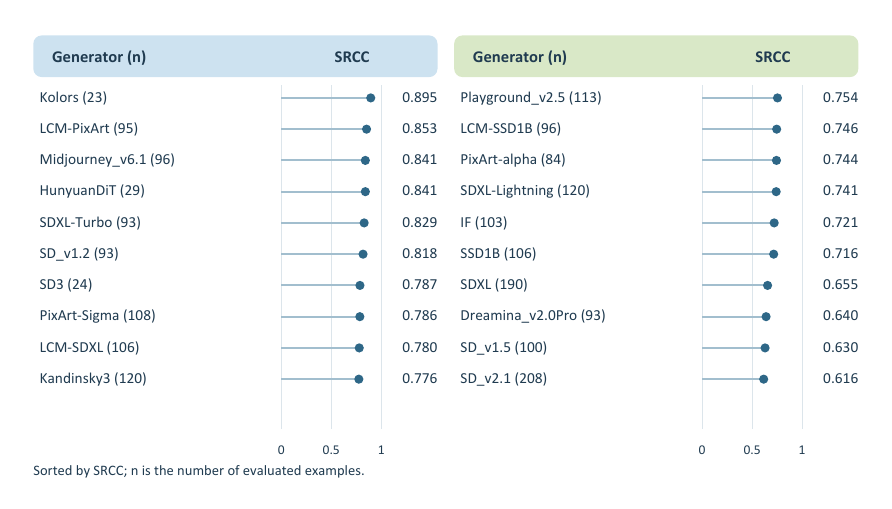}
\caption{
Per-generator SRCC of \score{}\ on the EvalMuse $2$K subset.
}
\label{fig:app-emgen}
\end{figure}

Per-generator score correlations on the external EvalMuse $2$K subset are
reported in Figure~\ref{fig:app-emgen}.

\paragraph{Prompt-family analysis.}
On the EvalMuse $2$K subset, SRCC is $0.7786$ on the $1{,}062$ examples
whose prompt families are absent from the EvalMuse training booster and
$0.7599$ on the $938$ examples whose prompt families appear in the booster.
The two subsets therefore show comparable score correlation.

\paragraph{EvalMuse subsets.}
Table~\ref{tab:app-evalmuse-splits} compares the primary-suite and external EvalMuse subsets. Official test labels are not public.

\begin{table}[!htbp]
\centering
\small
\setlength{\tabcolsep}{3pt}
\renewcommand{\arraystretch}{1.15}
\caption{Score correlation on two EvalMuse subsets.}
\begin{tabular*}{\textwidth}{@{\extracolsep{\fill}}lrcc@{}}
\toprule
EvalMuse slice & $n$ & \score{} SRCC & FGA-BLIP2 SRCC \\
\midrule
Primary suite & 200 & 0.803 & 0.918 \\
External 2K & 2{,}000 & 0.773 & 0.910 \\
\bottomrule
\end{tabular*}
\label{tab:app-evalmuse-splits}
\end{table}

\FloatBarrier
\subsection{Preference Evaluation on MMRB2}
\label{app:mmrb2}
We test whether the same scalar outputs support reward-model use on Multimodal RewardBench~2~\citep{hu2026multimodal}, using its text-to-image and image-editing subsets ($1{,}000$ expert-annotated preference pairs each). Candidates are scored independently, and the higher score determines the preference. Ties and unparseable pairs receive $0.5$, with all pairs retained per task. \score{} parses every pair, with $415$ T2I and $494$ editing ties. GPT-5.6-terra has $59$/$150$ unparseable pairs and $285$/$241$ ties on T2I/editing.

Qwen3-VL-8B uses the same evaluation instructions as \score{} without fine-tuning.

Among the $\chi_K$ evaluators in Table~\ref{tab:mmrb2}, \score{} ranks second in text-to-image accuracy ($0.574$), below GPT-5.6-terra ($0.614$) and above Qwen3-VL-8B ($0.566$). Its editing accuracy ($0.547$) is lower than that of GPT-5.6-terra ($0.619$) and Qwen3-VL-8B ($0.579$).

\begin{table}[!htbp]
\centering
\small
\setlength{\tabcolsep}{3pt}
\renewcommand{\arraystretch}{1.15}
\caption{MMRB2 preference accuracy ($1{,}000$ pairs per task). Rankings are separate for $\chi_0$ and $\chi_K$. Ties and unparseable pairs receive $0.5$.}

\begin{tabular*}{\textwidth}{@{\extracolsep{\fill}}llcc@{}}
\toprule
Method & Input & T2I & Edit \\
\midrule
ImageReward & $\chi_0$ & 0.530 & \underline{0.559} \\
PickScore & $\chi_0$ & \textbf{0.574} & \textbf{0.574} \\
HPSv2 & $\chi_0$ & 0.546 & 0.541 \\
VQAScore & $\chi_0$ & 0.537 & 0.554 \\
ImageDoctor & $\chi_0$ & 0.546 & 0.525 \\
Q-Align & $\chi_0$ & 0.513 & 0.523 \\
Q-Insight & $\chi_0$ & 0.492 & 0.521 \\
OmniQuality-R & $\chi_0$ & \underline{0.569} & 0.555 \\
\midrule
GPT-5.6-terra & $\chi_K$ & \textbf{0.614} & \textbf{0.619} \\
Qwen3-VL-8B judge & $\chi_K$ & 0.566 & \underline{0.579} \\
\score{} & $\chi_K$ & \underline{0.574} & 0.547 \\
\bottomrule
\end{tabular*}
\label{tab:mmrb2}
\end{table}

\FloatBarrier
\section{Ablation Studies}
\label{app:ablations}

\subsection{GRPO and Reward Design}
\label{app:grpo-studies}

\begin{figure}[!htbp]
\centering
\includegraphics[width=\textwidth]{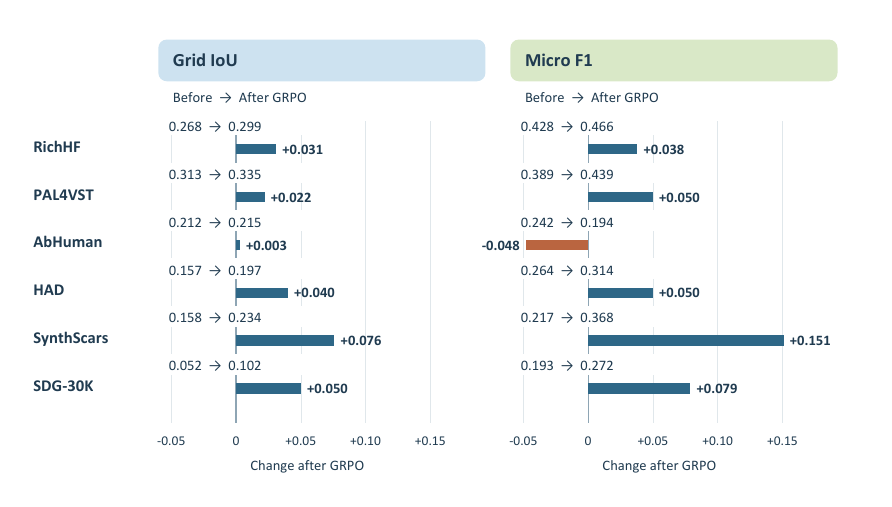}
\caption{Localization before and after GRPO. $\Delta$ is the final model minus \score{} ($-\mathrm{GRPO}$), computed from the displayed values.}
\label{fig:app-grpo-transfer}
\end{figure}

\paragraph{Transfer across benchmarks.}
\label{app:grpo-transfer}
Figure~\ref{fig:app-grpo-transfer} compares \score{} ($-\mathrm{GRPO}$) with the final model. GRPO improves per-image grid IoU on all six benchmarks and $F_1$ on five. On AbHuman, grid IoU rises slightly while pooled $F_1$ falls.

\FloatBarrier
\label{app:reward-seeds}

\paragraph{Reward ablation.}
Table~\ref{tab:rewarddecomp} compares the reward components under matched
training examples, sampling budget, and optimization steps. All variants use the same initialization and primary-suite evaluation support. We fix $\lambda_{\mathrm{dice}}=1$ and use weights $0.3$ and $0.1$ for included score and format rewards. $-r_{\mathrm{score}}$ and $-r_{\mathrm{format}}$ indicate that the corresponding reward weights are set to zero. Grid parse rate is $100\%$ for every variant.

\begin{table}[!htbp]
\centering
\small
\setlength{\tabcolsep}{3pt}
\renewcommand{\arraystretch}{1.15}
\caption{Reward ablation on the primary suite with matched initialization, data, sampling budget, and optimization steps.}
\label{tab:rewarddecomp}
\begin{tabular*}{\textwidth}{@{\extracolsep{\fill}}lccc@{}}
\toprule
Variant & Grid IoU & $F_1$ & Overall SRCC \\
\midrule
\score{} ($-\mathrm{GRPO}$) & 0.296 & 0.471 & \underline{0.596} \\
\score{} ($-r_{\mathrm{score}}, -r_{\mathrm{format}}$) & 0.320 & 0.481 & \textbf{0.601} \\
\score{} ($-r_{\mathrm{score}}$) & \underline{0.323} & \underline{0.496} & \underline{0.596} \\
\midrule
\score{} & \textbf{0.324} & \textbf{0.506} & \textbf{0.601} \\
\bottomrule
\end{tabular*}
\end{table}

\paragraph{Seed replication.}
All three GRPO runs improve both per-image grid IoU and $F_1$ over the shared SFT initialization (Table~\ref{tab:app-seeds}), with grid-IoU gains of $0.024$--$0.029$.

\begin{table}[!htbp]
\centering
\small
\setlength{\tabcolsep}{3pt}
\renewcommand{\arraystretch}{1.15}
\caption{Results across three GRPO runs and additional SFT. Changes are relative to SFT epoch 4 and computed before rounding.}
\begin{tabular*}{\textwidth}{@{\extracolsep{\fill}}lrrrr@{}}
\toprule
Run & $F_1$ & $\operatorname{IoU}^{\mathrm{grid}}_p$ & $\Delta F_1$ & $\Delta\operatorname{IoU}^{\mathrm{grid}}_p$ \\
\midrule
\score{} ($-\mathrm{GRPO}$) (epoch 4) & 0.471 & 0.296 & -- & -- \\
\score{} ($-\mathrm{GRPO}$) (epoch 5) & 0.472 & 0.297 & $+0.001$ & $+0.001$ \\
\score{} (seed 42) & 0.506 & 0.324 & $+0.034$ & $+0.028$ \\
\score{} (seed 2) & 0.483 & 0.320 & $+0.012$ & $+0.024$ \\
\score{} (seed 3) & 0.493 & 0.325 & $+0.022$ & $+0.029$ \\
\bottomrule
\end{tabular*}
\label{tab:app-seeds}
\label{tab:frontier}
\end{table}

\FloatBarrier
\paragraph{Precision--recall trade-off.}
\label{app:beta}
Increasing $\beta$ favors recall, while decreasing it favors precision. Table~\ref{tab:beta} compares the resulting development-set performance. We choose $\beta=1$ for its grid IoU; the Tversky variant gives higher $F_1$ but lower grid IoU. The Tversky reward~\citep{salehi2017tverskylossfunctionimage} is $\mathrm{TP}_w/(\mathrm{TP}_w+\alpha\,\mathrm{FP}+(1{-}\alpha)\,\mathrm{FN}_w)$, using the cell weights in Section~\ref{sec:rlvr}; $\alpha$ controls the false-positive penalty.

\begin{table}[!htbp]
\centering
\small
\setlength{\tabcolsep}{3pt}
\renewcommand{\arraystretch}{1.15}
\caption{Development-set precision--recall trade-offs. $\mathrm{FP}_{\mathrm{PAL}}/40$ counts false alarms on $40$ clean PAL4VST images.}
\begin{tabular*}{\textwidth}{@{\extracolsep{\fill}}lccccc@{}}
\toprule
Policy & $P$ & $R$ & $F_1$ & $\operatorname{IoU}^{\mathrm{grid}}_p$
& $\mathrm{FP}_{\mathrm{PAL}}/40$ \\
\midrule
\score{} ($-\mathrm{GRPO}$) & 0.374 & \underline{0.548} & 0.445 & 0.303 & \textbf{14} \\
GRPO $\beta=0.5$ & \textbf{0.452} & 0.450 & 0.451 & 0.300 & \underline{19} \\
GRPO $\beta=1$ & 0.411 & 0.534 & \underline{0.464} & \textbf{0.322} & 26 \\
GRPO $\beta=2$ & 0.334 & \textbf{0.695} & 0.451 & 0.310 & 27 \\
GRPO Tversky $\alpha{=}0.7$ & \underline{0.446} & 0.494 & \textbf{0.469} & \underline{0.313} & 21 \\
\bottomrule
\end{tabular*}
\label{tab:beta}
\end{table}

\FloatBarrier
\paragraph{Per-category accuracy and false alarms.}
\label{app:channel-clean}
Table~\ref{tab:app-channel-clean} separates per-category accuracy from clean-image false alarms. GRPO improves artifact F1 (0.488 to 0.493) and misalignment F1 (0.165 to 0.198), and raises both precision and recall on the primary suite (Table~\ref{tab:primary-full}). It also lowers coverage F1 (0.330 to 0.297) and increases false alarms on clean PAL4VST images (63 to 85 of 115), while COCO false alarms remain at zero. Misalignment localization remains weak in absolute terms for both models. These costs are consistent with the development-set trade-off in Table~\ref{tab:beta} and with the clean-image limitation noted in Section~\ref{sec:limitations}.

\begin{table}[!htbp]
\centering
\small
\setlength{\tabcolsep}{3pt}
\renewcommand{\arraystretch}{1.15}
\caption{Per-category localization on RichHF ($400$ examples) and clean-image false alarms. Coverage evaluates ``!'' marks.}
\begin{tabular*}{\textwidth}{@{\extracolsep{\fill}}lccccc@{}}
\toprule
Model & \shortstack{Artifact\\$F_1$} & \shortstack{Misalign.\\$F_1$} & \shortstack{Coverage\\$F_1$} & \shortstack{COCO\\FP/100} & \shortstack{PAL\\FP/115} \\
\midrule
\score{} ($-\mathrm{GRPO}$) & 0.488 & 0.165 & 0.330 & 0 & 63 \\
\score{} & 0.493 & 0.198 & 0.297 & 0 & 85 \\
SegFormer-b0 & -- & -- & -- & 34 & 62 \\
\bottomrule
\end{tabular*}
\label{tab:app-channel-clean}
\end{table}

\FloatBarrier
\subsection{Output Representation and Grid Resolution}
\label{app:representation}

\paragraph{Output representation.}
We compare several text-native spatial encodings using stand-alone
localizers trained with matched data and compute. Results are shown in
Table~\ref{tab:outputform}.

\begin{table}[!htbp]
\centering
\small
\setlength{\tabcolsep}{3pt}
\renewcommand{\arraystretch}{1.15}
\caption{Localization representations under matched training data and compute.}
\begin{tabular*}{\textwidth}{@{\extracolsep{\fill}}lcccc@{}}
\toprule
Output form & $P$ & $R$ & $F_1$ &
$\operatorname{IoU}^{\mathrm{grid}}_p$ \\
\midrule
Dense bitmap & \textbf{0.580} & 0.099 & 0.169 & 0.116 \\
Point list & 0.250 & 0.446 & 0.320 & 0.206 \\
Bounding boxes & \underline{0.354} & \textbf{0.671} & \textbf{0.463} & \textbf{0.280} \\
Sparse cells & 0.298 & \underline{0.596} & \underline{0.397} & \underline{0.246} \\
\bottomrule
\end{tabular*}
\label{tab:outputform}
\end{table}

Bounding boxes achieve higher precision, recall, $F_1$, and grid IoU than sparse cells in this stand-alone comparison. Nevertheless, we use sparse cells because the same coordinates support text generation, cell-level Dice rewards, and spatial explanations without an additional box-to-grid conversion. Cells can also represent disconnected regions and separate defect categories, although this experiment does not establish an accuracy advantage for those cases.

\paragraph{Grid resolution.}
Table~\ref{tab:gridchoice-full} examines the trade-off between spatial
granularity and autoregressive sequence length. The model-free columns
measure how well each grid preserves the original spatial annotation, while
the trained columns compare models trained under otherwise matched settings. Model-free metrics use $933$ non-empty annotations. In the table, ``GT vanished'' is the fraction of non-empty masks that become empty after grid conversion.

\begin{table*}[!htbp]
\centering
\small
\setlength{\tabcolsep}{3pt}
\renewcommand{\arraystretch}{1.15}
\caption{Grid resolution, annotation fidelity, and output length. Rankings apply to trained-model results.}
\begin{tabular*}{\textwidth}{@{\extracolsep{\fill}}rccccccc@{}}
\toprule
& \multicolumn{3}{c}{Representation ceiling (model-free)}
& \multicolumn{2}{c}{Serialized target}
& \multicolumn{2}{c}{Trained model} \\
\cmidrule(lr){2-4}
\cmidrule(lr){5-6}
\cmidrule(lr){7-8}
$N$ & pixel-F1 & pixel-IoU & \shortstack{GT\\vanished}
& \shortstack{tokens\\(mean)} & \shortstack{tokens\\(p95)}
& F1@$N$ & pixel-IoU$_p$ \\
\midrule
4  & 0.219 & 0.178 & 68.4\% & 7   & 35    & ---   & --- \\
8  & 0.477 & 0.374 & 24.1\% & 27  & 116   & \textbf{0.425} & 0.192 \\
12 & 0.657 & 0.535 & 9.8\%  & 66  & 250   & 0.398 & \textbf{0.235} \\
16
   & 0.731 & 0.615 & 6.4\%  & 120 & 471   & \underline{0.415} & \underline{0.231} \\
24 & 0.813 & 0.712 & 2.8\%  & 273 & 1{,}089 & 0.388 & 0.209 \\
32 & 0.857 & 0.770 & 1.5\%  & 486 & 1{,}926 & 0.366 & 0.192 \\
\bottomrule
\end{tabular*}
\label{tab:gridchoice-full}
\end{table*}

See Figure~\ref{fig:grid-tradeoff} in the main text for the resolution trade-off curves.

\FloatBarrier
\section{Qualitative Examples and Failure Analysis}
\label{app:qualitative}
We show held-out examples with ground-truth and predicted grids. In the walkthrough and task examples, red denotes artifacts, blue misalignment, and orange single-grid defects.

\paragraph{Selection for cross-benchmark comparisons.}
For Figure~\ref{fig:qualgrid}, candidate images have $6$--$70$ ground-truth defect cells. Within each benchmark, we select the image with the largest grid-IoU margin between \score{} and the highest-scoring competing prediction in the selection pool. We display the same four models on every example: \score{}, GPT-5.6-sol, Claude Opus 5.5, and ImageDoctor. Every panel retains the ground-truth outline, so a separate input column is unnecessary. This selection highlights favorable cases; aggregate comparisons are reported in Table~\ref{tab:spatialcompare}.

\subsection{Before and After GRPO}
\begin{figure}[t]
\centering
\includegraphics[width=\textwidth]{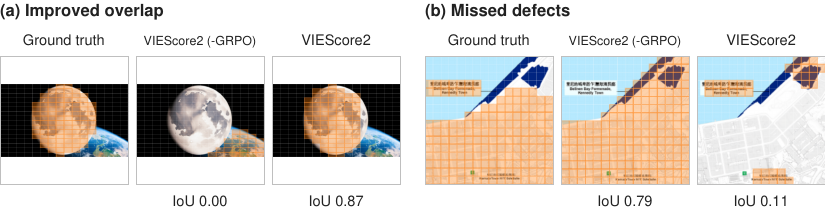}
\caption{Localization before and after GRPO. Orange cells mark defects; values are per-image grid IoU. \score{} ($-\mathrm{GRPO}$) uses the SFT checkpoint; epochs 4 and 5 give identical predictions. \score{} denotes the final model after GRPO.}
\label{fig:grpo-qualitative}
\label{fig:app-sft-rl}
\end{figure}

Figure~\ref{fig:grpo-qualitative} compares \score{} ($-\mathrm{GRPO}$) with the final \score{} model. In (a), GRPO shifts the predicted cells toward the annotated defect region, increasing grid IoU from $0.00$ to $0.87$. In (b), it misses much of the annotated region, reducing IoU from $0.79$ to $0.11$. An additional SFT epoch leaves both predictions unchanged. These examples illustrate the benefits and limitations of post-training; the aggregate gains do not imply improvement on every image.

\subsection{A Complete Evaluation Walkthrough}
\label{app:walkthrough}
\begin{figure}[!htbp]
\centering
\includegraphics[width=\textwidth]{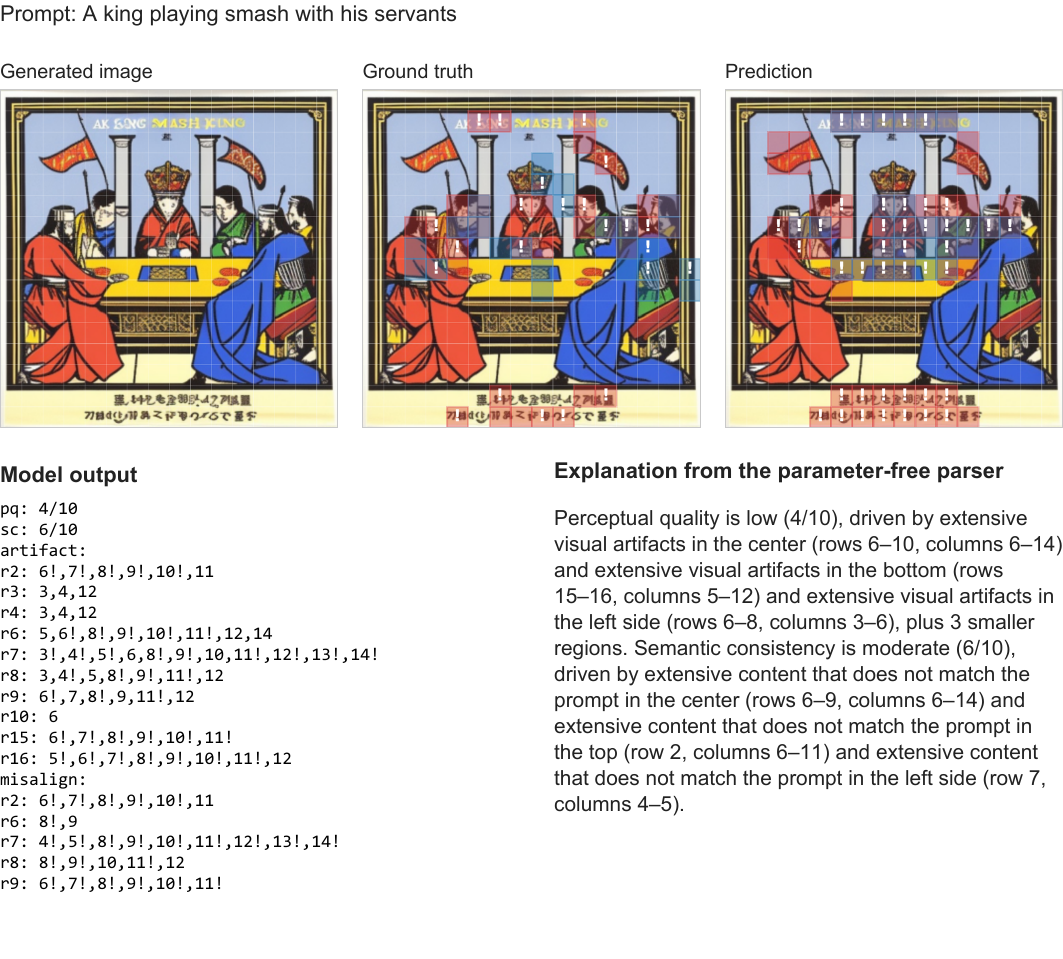}
\caption{A RichHF example with annotations, predictions, raw output, and the parameter-free parser's explanation.}
\label{fig:app-walkthrough}
\end{figure}

Figure~\ref{fig:app-walkthrough} shows a RichHF image, the complete model output, and the explanation from the parameter-free parser.

\clearpage
\subsection{Generation, Editing, and Reference-Conditioned Cases}
\label{app:task-examples}
\begin{figure}[!htbp]
\centering
\includegraphics[page=1,width=\textwidth]{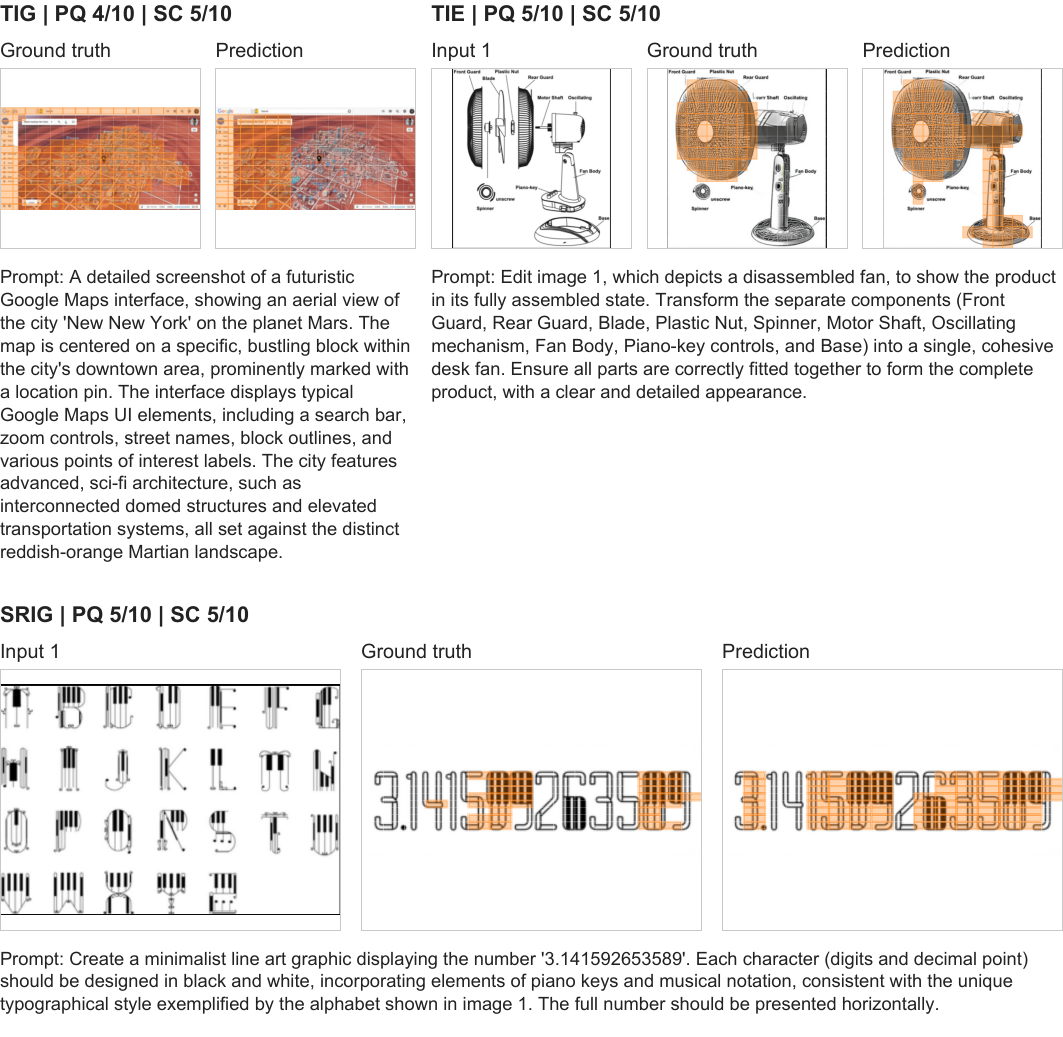}
\caption{TIG, TIE, and SRIG examples.}
\label{fig:app-task-gallery}
\end{figure}

\begin{figure}[!htbp]
\centering
\includegraphics[page=2,width=\textwidth]{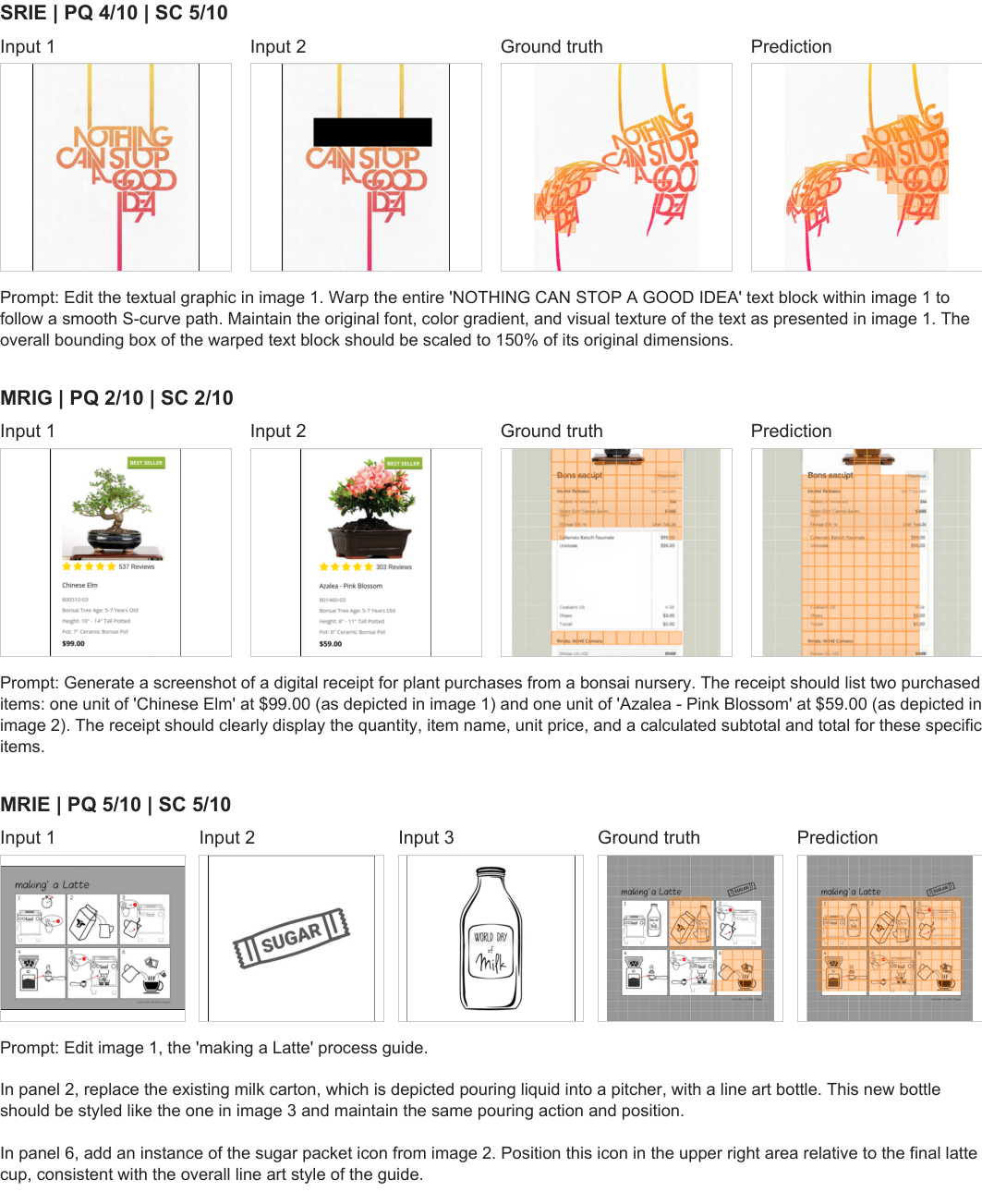}
\caption{SRIE, MRIG, and MRIE examples.}
\label{fig:app-reference-gallery}
\end{figure}

Figures~\ref{fig:app-task-gallery} and~\ref{fig:app-reference-gallery} show the six ImagenWorld tasks with their inputs and original prompts. Panels within each task row use matched frames.

\clearpage

\clearpage
\subsection{Failure Cases and Interpretation Limits}
\label{app:failures}
\begin{figure}[!htbp]
\centering
\includegraphics[width=\textwidth]{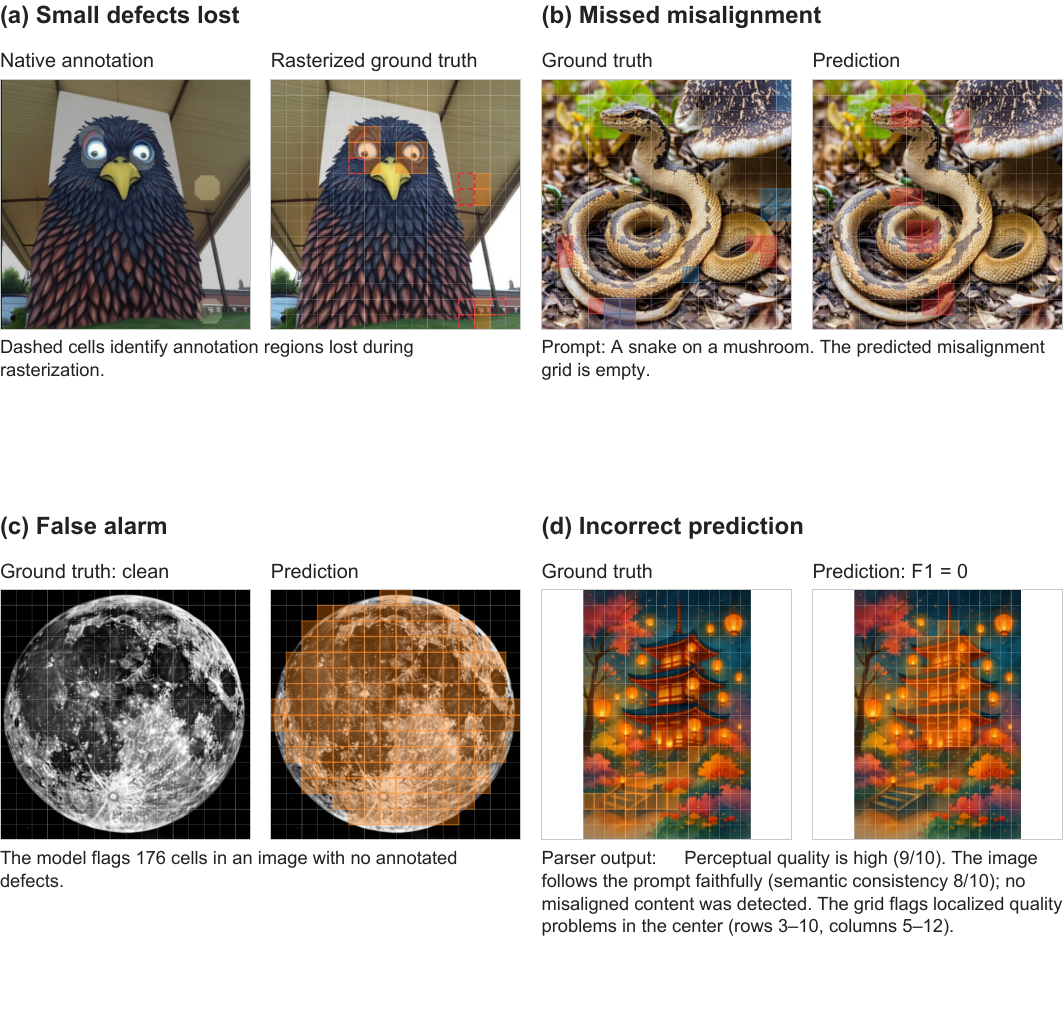}
\caption{Failure cases: lost small defects, missed misalignment, false alarms, and an incorrect prediction with its explanation.}
\label{fig:app-failures}
\end{figure}

Figure~\ref{fig:app-failures} illustrates representation loss, missed misalignment, false alarms, and an explanation generated from an incorrect prediction.

\clearpage
\section{Evaluation Instruction Templates}
\label{app:prompts}

The evaluation instructions below specify image order, requested fields, and output syntax. \texttt{<PROMPT>} denotes the generation or editing prompt. In the original wording, ``channels'' refers to defect categories, and ``severe'' or ``severely'' accompanies ``!''; its supervision is annotation coverage (Appendix~\ref{app:heatmap-processing}).


\begin{promptbox}{Shared system message}
You are an expert image quality evaluator.

You may be shown one or more INPUT images --- an original image to be edited and/or reference images that were given to the image generator --- followed by the GENERATED image to be evaluated. The GENERATED image is ALWAYS the LAST image; earlier images are referred to as "image 1", "image 2", etc. in reading order.
When input images are present, judge the generated image RELATIVE to them: whether the requested edit was applied correctly, whether content that should have been preserved was preserved, and whether referenced subjects are reproduced faithfully. When no input image is present, judge the generated image against the text prompt alone.

The GENERATED image is divided into a 16\ensuremath{\times}16 grid of cells (row 1 is the top, column 1 is the left).
For every cell of the GENERATED image decide whether it contains visible quality problems: artifacts, distortions, blurriness, missing or malformed content, color abnormalities, unrealistic content, or --- for edits --- a failed/incorrect edit or a wrongly altered region.

FIRST output one line with the overall quality score of the generated image in the form "score: \textless{}0-10\textgreater{}/10" (10 = flawless, 0 = unusable).
THEN list ONLY the cells that HAVE problems, one line per affected row, in the form "r\textless{}row\textgreater{}: \textless{}col\textgreater{},\textless{}col\textgreater{},..." using 1-based indices (rows and columns range from 1 to 16). Emit a line only for rows that contain at least one problematic cell, in increasing row order. If the image has no problems at all, output exactly "none" instead of the cell lines.

When the request asks for separate problem CHANNELS, report the cells in two sections instead of one list: a header line "artifact:" followed by the rows of cells with visual artifacts (distortions, malformed or implausible content, rendering errors), then a header line "misalign:" followed by the rows of cells whose content contradicts the prompt (wrong, missing or extra objects/attributes). Within each section use the same "r\textless{}row\textgreater{}: \textless{}col\textgreater{},..." lines, appending "!" to a column number when that cell's problem is severe (e.g. "r5: 8,9!,10"). Write "none" under a section with no problems.

When the request asks for TWO scores instead of one, rate perceptual quality (freedom from artifacts/distortions) as "pq: \textless{}0-10\textgreater{}/10" and semantic consistency with the prompt as "sc: \textless{}0-10\textgreater{}/10", one line each, before the cell lines.
\end{promptbox}

\clearpage
\begin{promptbox}{RichHF user template}
This generated image is divided into a 16\ensuremath{\times}16 grid of equal cells.\par
Generation/edit prompt: "\textless{}PROMPT\textgreater{}"\par
First rate the GENERATED image on two axes --- perceptual quality (artifacts, distortions, rendering flaws) as "pq: \textless{}0-10\textgreater{}/10" and semantic consistency with the prompt as "sc: \textless{}0-10\textgreater{}/10", one line each --- then list the problem cells of the GENERATED image in two channels: an "artifact:" section for cells with visual artifacts, then a "misalign:" section for cells whose content contradicts the prompt. In each section use one line per affected row as "r\textless{}row\textgreater{}: \textless{}cols\textgreater{}", append "!" to severely affected columns, and write "none" under an empty section.
\end{promptbox}

\begin{promptbox}{Conditional user template (two input images)}
The first 2 images (image 1, image 2) are the input/reference images that were given to the generator. The LAST image is the GENERATED image to evaluate; it is divided into a 16\ensuremath{\times}16 grid of equal cells.\par
Generation/edit prompt: "\textless{}PROMPT\textgreater{}"\par
First rate the GENERATED image on two axes --- perceptual quality (artifacts, distortions, rendering flaws) as "pq: \textless{}0-10\textgreater{}/10" and semantic consistency with the prompt as "sc: \textless{}0-10\textgreater{}/10", one line each --- then list the cells of the GENERATED image that contain visible quality problems, one line per affected row as "r\textless{}row\textgreater{}: \textless{}cols\textgreater{}". Output "none" instead of cell lines if there are no problems.
\end{promptbox}

\begin{promptbox}{Localization-only user template}
This generated image is divided into a 16\ensuremath{\times}16 grid of equal cells.\par
Generation/edit prompt: "\textless{}PROMPT\textgreater{}"\par
List the cells of the GENERATED image that contain visible quality problems, one line per affected row as "r\textless{}row\textgreater{}: \textless{}cols\textgreater{}". Output "none" instead of cell lines if there are no problems.
\end{promptbox}

\begin{promptbox}{EvalMuse score-only user template}
Generation prompt: "\textless{}PROMPT\textgreater{}"\par
Give the overall quality score of the GENERATED image as "score: \textless{}0-10\textgreater{}/10".
\end{promptbox}

\end{document}